# MambaX-Net: Dual-input Mamba-enhanced Cross-Attention network for longitudinal prostate MRI segmentation

Yovin Yahathugoda [a,*], Davide Prezzi [a,b], Patricia A. Gutierrez [a,c], Piyalitt Ittichaiwong [a], Vicky Goh [a,b], Sebastien Ourselin [a], Michela Antonelli [a]

[a] School of Biomedical Engineering & Imaging Sciences, King's College London, United Kingdom
[b] Department of Radiology, Guy's and St Thomas' NHS Foundation Trust, United Kingdom
[c] Department of Radiology, Centre Hospitalier Dunkerque, F-59240 Dunkirk, France

ABSTRACT

Active Surveillance (AS) is a treatment option for managing low and intermediate-risk prostate cancer (PCa), aiming to avoid overtreatment while monitoring disease progression through serial MRI and clinical follow-up. Accurate prostate segmentation is an important preliminary step for automating this process, enabling automated detection and diagnosis of PCa. However, existing deep-learning segmentation models are often trained on single-time-point, expertly annotated datasets, making them unsuitable for longitudinal AS analysis, where multiple time points and a scarcity of expert labels hinder effective fine-tuning. To address these challenges, we propose MambaX-Net, a novel semi-supervised, dual-scan 3D segmentation architecture that computes the segmentation for time point $t$ by leveraging the MRI and the corresponding segmentation mask from the previous time point. We introduce two new components: (i) a Mamba-enhanced Cross-Attention Module, which integrates the Mamba block into cross-attention to efficiently capture temporal evolution and long-range spatial dependencies, and (ii) a Shape Extractor Module that encodes the previous segmentation mask into a latent anatomical representation for refined zone delineation. Moreover, we use a semi-supervised self-training strategy that leverages pseudo-labels generated from a pre-trained nnU-Net, enabling effective learning without expert annotations. MambaX-Net was evaluated on a longitudinal AS dataset, and results showed that it significantly outperforms state-of-the-art U-Net and Transformer-based models, achieving superior prostate zone segmentation even when trained on limited and noisy data.

## 1. Introduction

Prostate cancer (PCa) is one of the most prevalent cancers among men in Western countries (Chen et al., 2013). The majority of patients diagnosed with PCa present with low-intermediate-risk disease (International Society of Urological Pathology, ISUP grade 1 and 2), localised to the prostate (Hamdy et al., 2023). Active surveillance (AS) is a treatment option for low-intermediate-risk PCa, which limits unnecessary treatments and their adverse effects (Ploussard et al., 2022) while monitoring disease progression. Current AS protocols require regular clinical examinations, serum prostate-specific antigen (PSA) testing, repeated Magnetic Resonance Imaging (MRI) and biopsy to confirm progression (Hamdy et al., 2023). Although MRI has emerged as an important non-invasive tool for selecting and monitoring patients in AS (Briganti et al., 2018), its interpretation requires substantial expertise and remains time-intensive, particularly for complex AS cases (Kania et al., 2025), making Artificial Intelligence (AI) a promising approach to automate and streamline prostate MRI analysis, improving diagnostic efficiency (Arvidsson et al., 2024).

A critical pre-requisite for AI-driven prostate MRI analysis is accurate segmentation of the whole prostate (WP) as well as its sub-regions, particularly the peripheral (PZ) and transition (TZ) zones, with the different zones having distinct pathogenetic features of PCa (Yu et al., 2023). Accurate WP segmentation is essential for estimating prostate volume, a key requirement for calculating PSA density (PSAd), which helps distinguish elevated PSA caused by PCa from benign prostatic hyperplasia (Kania et al., 2025), while PZ and TZ segmentations are crucial priors for automated lesion detection (Zheng et al., 2024).

* Corresponding author.
*E-mail addresses:* yovin.yahathugoda@kcl.ac.uk (Y. Yahathugoda), davide.prezzi@kcl.ac.uk (D. Prezzi), patricia.gutierrez@kcl.ac.uk (P.A. Gutierrez), piyalitt.ittichaiwong@kcl.ac.uk (P. Ittichaiwong), vicky.goh@kcl.ac.uk (V. Goh), sebastien.ourselin@kcl.ac.uk (S. Ourselin), michela.antonelli@kcl.ac.uk (M. Antonelli).

Prostate zone segmentation approaches typically are based on either Convolutional Neural Networks (CNNs) or Transformers. While the CNN-based nnU-Net (Isensee et al., 2021) remains state-of-the-art (SOTA) and a powerful baseline (Bhandary et al., 2023; Rodrigues et al., 2023), Transformer-based models such as SwinUNETR (He et al., 2023) are increasingly explored for their ability to capture long-range global context. Yet, Transformers often struggle to preserve fine-grained details and typically require large training datasets to surpass CNNs (Yan et al., 2023). Yan et al. 2023 reported that their hybrid CNN-ViT model, CCT-UNet, outperformed U-Net and SwinUNETR, though its Transformer backbone relied on a large training dataset. Huang et al. 2024 showed that the U-Net was superior to a ViT in learning effectively from a small labelled dataset under semi-supervised settings. The findings suggest that while attention mechanisms may enhance CNNs, simple integration is insufficient; for example, a Global–Local Channel and Spatial Attention-enhanced U-Net (Krishnan et al., 2025) was outperformed by the standard nnU-Net baseline used in our experiments on the ProstateX dataset, suggesting that more advanced approaches are required to improve upon highly-optimised architectures.

Recent advances in prostate MRI segmentation have addressed two interrelated challenges: improving zone boundary delineation and reducing annotation burden using semi-supervised approaches. In terms of architecture, multi-scale context fusion approaches such as ProSeg (Qin and Yang, 2026) have improved boundary precision by integrating hierarchical feature representations. Deng et al. 2026 proposed a cooperative morphology boundary network to address ambiguous boundaries between the prostate and surrounding tissues, as well as substantial inter-patient shape variability. To reduce reliance on manual annotations, semi-supervised approaches such as entropy minimisation, cross pseudo-supervision, mean teacher, uncertainty-aware mean teacher (UAMT), and interpolation consistency training (ICT) have been investigated for prostate segmentation (Huang et al., 2024). Another strategy is to use a single expert-annotated mid-slice and apply semi-supervised learning to extrapolate the remaining slices from the sparse annotations (Osman et al., 2024). However, the authors reported degraded segmentation performance at the top and bottom slices, compared with the central slice of the prostate. Crucially, existing prostate segmentation studies are based on single-time-point imaging, while longitudinal prostate MRI studies focus mainly on lesion progression or outcome prediction rather than prostate segmentation.

Together, the architectural limitations and the unmet annotation challenges underscore the need for novel architectures that can overcome the limitations of both CNNs and Transformers. Recently, the Mamba architecture (Gu and Dao, 2024) has gained popularity in medical imaging. Unlike Transformers that rely on a computationally intensive self-attention mechanism scaling quadratically with sequence length, Mamba scales linearly, offering significant advantages over Transformers, especially for 3D modalities like MRI. At its core, Mamba blocks use a selection mechanism that makes its parameters input-dependent to allow content-based reasoning. This allows the model to dynamically decide which information to keep and which to forget, effectively compressing relevant context into a fixed-size state and capturing long-range spatial dependencies. The selection mechanism in Mamba is enhanced by a hardware-aware algorithm that uses the parallel scan to run the recurrent computation in parallel on GPUs, combining the efficiency of recurrent models with the content-aware reasoning of Transformers. Recent works have integrated Mamba into hybrid architectures, including SegMamba (Xing et al., 2024), MUNet (Yang et al., 2025), and MambaBTS (Zhou et al., 2024), which combine the powerful local feature extraction of CNNs with Mamba's efficient global context modelling, leading to superior performance in medical image segmentation. Recent Mamba-based hybrids have also explored the combination of state-space models and attention, including CrossMamba (Wu et al., 2025) for audio source extraction, BiCrossMamba-ST (Kheir et al., 2025) for spectro-temporal speech deepfake detection, and TM-Unet (Xu and Yin, 2025) for single-time-point medical image segmentation.

While Mamba-based architectures have achieved competitive performance in 3D segmentation tasks, they, like other SOTA prostate segmentation models (Fassia et al., 2024), are exclusively designed for single-time-point data and have not been extended to longitudinal analysis, which is crucial in the context of AS. Single-time-point models are inherently limited for longitudinal monitoring as they treat each scan in isolation, failing to leverage the rich temporal information available across patients' imaging history and to capture the underlying anatomical changes. In other domains, such as brain imaging, SOTA longitudinal segmentation approaches typically rely on registering scans across different time points (Andrearczyk et al., 2024; Cerri et al., 2023). However, these methods are often unreliable, as they are prone to registration errors, especially for prostate segmentation, where substantial non-rigid deformation over time makes registration challenging. In our prior work, we introduced the Dual-Scan Model (DSM) (Yahathugoda et al., 2025), which was the first attempt to incorporate longitudinal information for prostate MRI segmentation by jointly processing two subsequent time points. However, DSM relied on cross-attention combined with SimAM blocks (Yang et al., 2021), which is effective for local spatial suppression but cannot effectively capture long-range temporal dependencies or condition the model on the previous prostate-zone mask as an anatomical prior, often producing irregular segmentations at the prostate apex.

In addition to architectural design, model performance also depends on the availability of expertly annotated datasets, which are particularly limited in AS due to the time-consuming task of labelling multiple time points per patient (Becker et al., 2019; Kitamura et al., 2024). Pre-training on public datasets followed by fine-tuning on the target domain dataset is a promising strategy (Alzate-Grisales et al., 2023; Chae and Kim, 2023; Guan and Liu, 2022), but fine-tuning on the target domain dataset is still dependent on the availability of expert annotations, which can be scarce for longitudinal datasets.

To address both the lack of longitudinal modelling and the scarcity of expert-annotated longitudinal data, we introduce MambaX-Net, a novel 3D segmentation model designed specifically for longitudinal data, together with a self-training framework that leverages pseudo-labels to reduce annotation requirements. MambaX-Net is a dual-scan architecture that extends a 3D nnU-Net to segment the image at time point $t$ using information from the image and the corresponding segmentation mask at time point $t-1$. Its novelty lies in two key aspects: (i) the Mamba-enhanced Cross-Attention Module (M-CAM), which integrates Mamba blocks into cross-attention (Mital et al., 2023) to capture temporal dependencies, and (ii) the injection of the segmentation mask from the previous time point through the Shape Extractor Module (SEM), a sequential block of convolutional layers with LeakyReLU activations that encodes the prostate zone mask into a latent representation of key anatomical features.

Compared with our previous DSM (Yahathugoda et al., 2025), MambaX-Net extends the dual-scan paradigm in two complementary directions: M-CAM provides long-range temporal modelling beyond the local spatial suppression of SimAM, while SEM introduces explicit prior-mask conditioning that was absent from DSM. The complementarity of M-CAM and SEM is central to this design. M-CAM performs global inter-scan feature alignment, but without an explicit structural anchor, it may remain sensitive to non-rigid anatomical changes that shift apparent zone boundaries. Conversely, SEM provides this patient-specific anatomical anchor by encoding the previous segmentation mask, but its shallow architecture alone cannot resolve long-range dependencies across the gland, particularly at the apex and base. By injecting the compact SEM representation directly into M-CAM, MambaX-Net grounds Mamba's global attention modelling in patient-specific geometry. To the best of our knowledge, MambaX-Net is the first Mamba-based segmentation network to jointly model longitudinal MRI evolution and explicit prior-mask anatomical context.

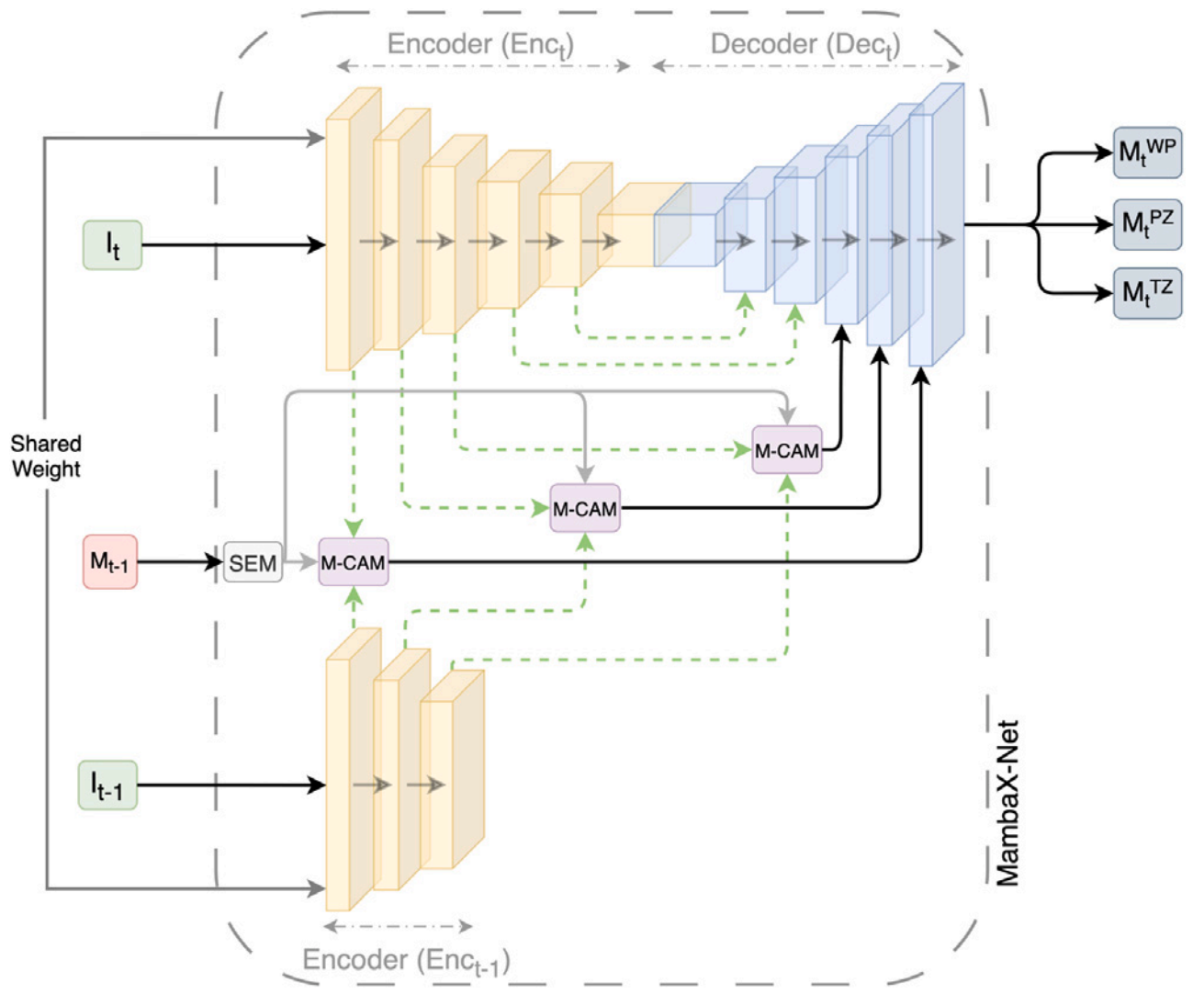

**Fig. 1.** High-level architecture of MambaX-Net showing a U-Net-like dual-encoder structure ($Enc_t$, $Enc_{t-1}$) with shared weights for processing scans at two time points ($I_t$ and $I_{t-1}$), and a single decoder ($Dec_t$). The M-CAM block fuses latent features from $I_t$ and $I_{t-1}$ with the SEM output, producing enhanced representations for prostate zone segmentation.

Regarding the self-training framework, pseudo-labels for all the AS timepoints are generated using an nnU-Net model trained on a public dataset. The pseudo-label at $t-1$ is provided to MambaX-Net during both training and inference for segmentation refinement, whereas the pseudo-label at $t$ is used only for loss calculation during fine-tuning.

We evaluated MambaX-Net against three SOTA baselines (nnU-Net-V2, SwinUNETR-V2, SegMamba) as well as the DSM. Experimental results show that MambaX-Net significantly outperforms all four models, generating more accurate prostate zone segmentations even when trained on a reduced dataset with noisy labels. This performance gain derives from MambaX-Net's ability to capture global anatomical relationships through M-CAM and refine segmentation boundaries with SEM, resulting in more accurate and robust prostate zone segmentation, particularly in data-limited settings.

## 2. Methods

This section presents the architecture of our novel MambaX-Net model, its core components, and the loss function used to train the model.

### 2.1. MambaX-Net architecture overview

The MambaX-Net architecture shown in Fig. 1 extends the 3D nnU-Net encoder–decoder architecture, originally designed for single-timepoint data, with key architectural improvements that enable the integration of longitudinal information of the same patient from two sequential MRI scans (time point $t$ and $t-1$) and the prior segmentation mask (time point $t-1$).

For each patient, let $I_t$ denote the image to be segmented at time point $t$, $I_{t-1}$ the image acquired at time point $t-1$, and $M_{t-1}$ its corresponding 3-channel pseudo-label representing the whole prostate (WP), peripheral zone (PZ), and transition zone (TZ). MambaX-Net employs dual encoders, $Enc_t$ and $Enc_{t-1}$, with shared weights to process $I_t$ and $I_{t-1}$, respectively, together with a decoder, $Dec_t$, analogous to a standard U-Net architecture. $Enc_t$ consists of six downsampling blocks to extract multi-scale feature maps ($f_t^i, i \in \{1,\ldots,6\}$) from $I_t$, while the auxiliary encoder $Enc_{t-1}$ is truncated to extract only the top three feature maps ($f_{t-1}^i$, $i \in \{1,\ldots,3\}$) from $I_{t-1}$. This design choice was determined empirically during hyperparameter optimisation. Limiting $Enc_{t-1}$ to these feature levels reduces computational complexity while preserving an optimal trade-off between high-resolution spatial details from shallower layers and rich semantic context from deeper layers.

In parallel, the Shape Extraction Module (SEM) shown in Fig. 1, which comprises three 3D convolutional blocks, encodes $M_{t-1}$ into a latent representation $f_{t-1}^{SEM}$ capturing key anatomical features, such as the size and shape of the prostate. M-CAM then takes as input $f_t^i$, and $f_{t-1}^i$ combined with $f_{t-1}^{SEM}$ to perform implicit image registration in the feature space across $I_t$, $I_{t-1}$, and $M_{t-1}$, and integrate longitudinal information into a latent representation.

The fused features from M-CAM are injected into the last three upsampling blocks of the decoder, whereas the earlier blocks rely solely on standard skip connections. The model outputs a three-channel 3D segmentation mask ($M_t$) comprising WP, PZ, and TZ.

### 2.2. SEM and M-CAM for latent feature fusion

The proposed Mamba-enhanced Cross-Attention Module (M-CAM), shown in Fig. 2, leverages Mamba blocks (Gu and Dao, 2024) to efficiently model long-range dependencies with linear scaling, providing robust performance across training datasets of varying sizes. M-CAM is the core architectural component of MambaX-Net, designed to align and fuse feature maps $f_t^i$, $f_{t-1}^i$ and $f_{t-1}^{SEM}$.

The design of SEM and M-CAM is motivated by the anatomical consistency and temporal continuity observed in prostate gland shape and zonal anatomy on MRI during active surveillance. The WP, PZ and TZ have patient-specific but spatially constrained morphology; although image intensity, scan quality, and gland volume may vary between visits, the overall zonal topology usually changes gradually. SEM therefore encodes the previous 3-channel prostate-zone mask as an anatomical prior using three 3D convolutional blocks, each comprising a convolutional block consisting of a 3D convolution layer, an instance normalisation layer and a LeakyReLU activation layer. All three blocks use a stride of 1, preserving the full spatial resolution of the input mask throughout. The three blocks progressively map the 3-channel binary zone mask to 16, 32, and then 64 feature channels, encoding the prior prostate anatomy. M-CAM fuses this structural prior with longitudinal image features from the previous scan ($f_{t-1}^i$), which serve as anatomically informed keys and values, and the current scan features ($f_t^i$) as queries for cross-attention. The proposed design allows the model to use temporally consistent prostate anatomy as contextual guidance, with SEM encoding the prior prostate-zone mask as an anatomical shape prior and M-CAM aligning features from the previous and current MRI scans, while preserving current scan information through residual feature fusion. By encoding the previous mask as a latent prior and fusing it within M-CAM, the model is guided by prior anatomical information while still relying on current-scan image features for its final prediction. This architectural design prevents the network from learning a trivial shortcut, namely copying the previous segmentation masks directly to the output rather than segmenting the current scan. This approach effectively helps refine weak or ambiguous prostate zone boundaries. In our implementation, M-CAM is applied to the first three encoder feature levels, with 256-dimensional token embeddings, four attention heads, and a dropout of 0.2. The Mamba blocks use the default configuration from Gu and Dao 2024.

First, $f_t^i$ and $f_{t-1}^i$ are processed by a patch embedding layer, consisting of a linear layer and a rearrangement operation, followed by a normalisation layer. This yields the 3D patch embeddings $P_t^i$ and $P_{t-1}^i \in \mathbb{R}^{B\times N\times E}$, where $B$ is the batch size, $N$ is the number of patches, and $E$ is the embedding dimension.

Because the SEM output and encoder features may have different spatial resolutions and channel dimensions, the SEM feature map is first resized to match the spatial resolution of the previous-scan feature map $f_{t-1}^i$. It is then projected with a $1\times1\times1$ convolution to match

**Algorithm 1** SEM-guided M-CAM feature fusion

**Input:** Current MRI $I_t$, previous MRI $I_{t-1}$, previous prostate-zone mask $M_{t-1}$

**Output:** Predicted current prostate-zone mask $\hat{M}_t$

**Steps:**

$\{f_t^i\} \leftarrow \mathrm{Enc}(I_t)$

$\{f_{t-1}^i\} \leftarrow \mathrm{Enc}(I_{t-1})$

$f_{t-1}^{\mathrm{SEM}} \leftarrow \mathrm{SEM}(M_{t-1})$

**for** each selected M-CAM level $i$ **do**

  $P_t^i \leftarrow \mathrm{PatchEmbed}(f_t^i)$

  $P_{t-1}^i \leftarrow \mathrm{PatchEmbed}(f_{t-1}^i)$

  $s_{t-1}^i \leftarrow \mathrm{LNorm}(\mathrm{Proj}(\mathrm{Resize}(f_{t-1}^{\mathrm{SEM}})))$

  $S_{t-1}^i \leftarrow \mathrm{PatchEmbed}(s_{t-1}^i)$

  $\hat{P}_{t-1}^i \leftarrow P_{t-1}^i + S_{t-1}^i$

  $Z_t^i \leftarrow \mathrm{Mamba}(\mathrm{LNorm}(P_t^i))$

  $Z_{t-1}^i \leftarrow \mathrm{Mamba}(\mathrm{LNorm}(\hat{P}_{t-1}^i))$

  $A_t^i \leftarrow \mathrm{CrossAttn}(Q{=}Z_t^i,\ K{=}Z_{t-1}^i,\ V{=}Z_{t-1}^i)$

  $f_{att}^i \leftarrow \mathrm{Unpack}(A_t^i)$

  $f_{\mathrm{CAM},t}^i \leftarrow f_t^i + f_{att}^i$

**end for**

**return** $\hat{M}_t \leftarrow \mathrm{Dec}(\{f_{\mathrm{CAM},t}^i\})$

the channel dimension of the corresponding encoder level. The projected SEM feature map is passed through the same patch embedding operation as the image features, producing level-specific SEM patch embeddings $S_{t-1}^i \in \mathbb{R}^{B\times N_i\times E}$. These SEM patch embeddings are added directly to the previous scan patch embeddings $P_{t-1}^i$ to obtain the SEM-enhanced representation, injecting anatomical context after patch embedding while keeping the current and previous-scan embedding spaces aligned:

$$\hat{P}_{t-1}^i = P_{t-1}^i + S_{t-1}^i.$$

Both patch embeddings, $P_t^i$ and $\hat{P}_{t-1}^i$, are then normalised and processed by separate Mamba blocks. As a result, Mamba operates on image-derived patch embedding sequences rather than directly on the original 5D feature maps, enabling long-range dependencies to be modelled across the flattened 3D patch representation. The Mamba-enhanced current patch embeddings are used as queries, while the Mamba-enhanced previous patch embeddings provide keys and values for cross-attention. The attended patch embedding representation is subsequently passed through an unpacking operation that restores the spatial feature map layout, producing $f_{att}^i$. Finally, residual fusion is performed by adding $f_{att}^i$ to the original current-scan feature map:

$$f_{CAM,t}^i = f_t^i + f_{att}^i.$$

This residual addition preserves the dimensionality of $f_t^i$, while allowing temporally consistent anatomical information from $f_{t-1}^i$ and $f_{t-1}^{SEM}$ to guide feature fusion.

A key design choice in integrating SEM with M-CAM is to inject $f_{t-1}^{SEM}$ after the patch embedding layer rather than before. The patch embedding layer projects semantically rich feature maps into a sequence of patches, effectively translating the spatial grid into a format suitable for Mamba and cross-attention. Injecting $f_{t-1}^{SEM}$ before the patch embedding layer would create patches from a composite feature map, where the semantic content of $f_{t-1}^i$ could be distorted by structural details from $f_{t-1}^{SEM}$, leading to less distinct and meaningful representations. By contrast, adding $f_{t-1}^{SEM}$ after the patch embedding layer allows structural information to refine and guide the semantic representations in $P_{t-1}^i$, without disrupting their encoding. This design also makes M-CAM modular: if a prior segmentation mask is unavailable, the SEM branch can simply be omitted. However, removing $f_{t-1}^{SEM}$ eliminates its semantic enhancement and negatively impacts the overall model performance.

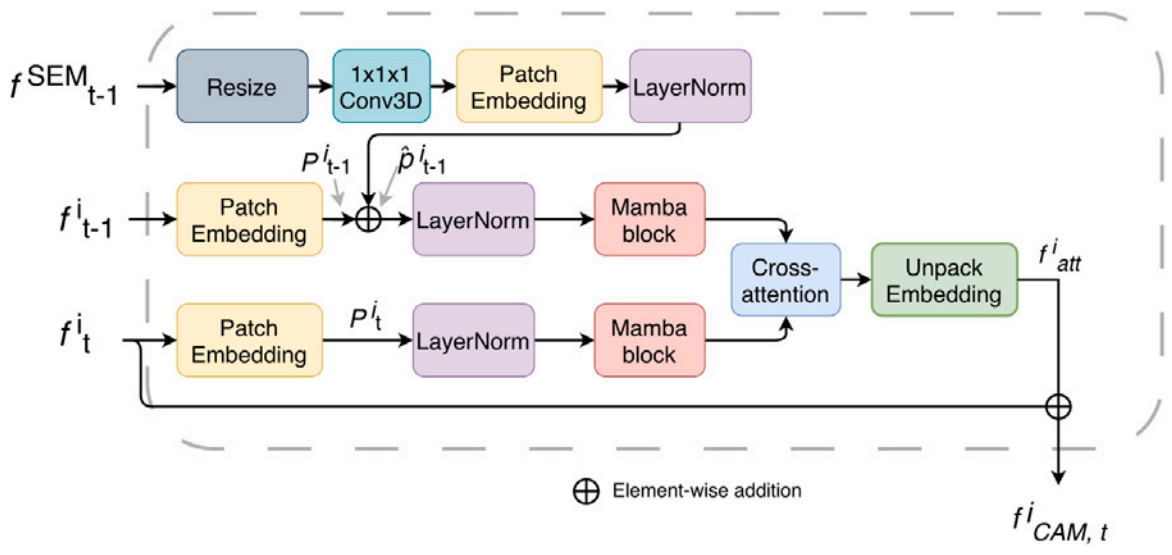

**Fig. 2.** The Mamba-enhanced Cross-Attention Module (M-CAM) structure.

### 2.3. Loss function

Designing an effective loss function is crucial for accurate prostate segmentation, particularly at the apex and base, where the gland is small and anatomically irregular. Previous research has shown that dice-cross-entropy (DiceCE) loss struggles with segmenting irregular regions, while the use of the Focal and Tversky loss functions enables the loss to focus on hard negative examples, thereby reducing false positives and false negatives (Rodrigues et al., 2023). For this reason, we employed a modified version of the Focal Tversky Dice Cross-Entropy loss (FT_DCE), shown in Eq. (1), where instead of averaging the three loss components (Focal, Dice, and Tversky), we combined the Focal and Tversky losses (FT) following (Abraham and Khan, 2019), and incorporated them with the DiceCE loss (Isensee et al., 2021) using a weighted sum, with the weight parameter $\omega$ set empirically through hyperparameter optimisation.

$$\mathcal{L}_{(\mathrm{FT_DCE})} = \omega\mathcal{L}_{\mathrm{DiceCE}} + (1-\omega)\mathcal{L}_{\mathrm{FT}} \tag{1}$$

$\mathcal{L}_{\mathrm{DiceCE}}$ is expressed in Eq. (2) with $\mathcal{L}_{\mathrm{Dice}}$ and $\mathcal{L}_{\mathrm{BCE}}$ defined by Eq. (3) and (4), respectively.

$$\mathcal{L}_{\mathrm{DiceCE}} = \mathcal{L}_{\mathrm{Dice}} + \mathcal{L}_{\mathrm{BCE}} \tag{2}$$

$$\mathcal{L}_{\mathrm{Dice}} = 1 - \frac{2\sum_{i=1}^{N} p_i g_i}{\sum_{i=1}^{N} p_i^2 + \sum_{i=1}^{N} g_i^2} \tag{3}$$

$$\mathcal{L}_{\mathrm{BCE}} = \begin{cases} -\log(p_i) & \text{if the true label } g_i \text{ is } 1 \\ -\log(1-p_i) & \text{if the true label } g_i \text{ is } 0 \end{cases} \tag{4}$$

where for each voxel $i$, $p_i \in [0,1]$ is the predicted probability and $g_i \in \{0,1\}$ the ground-truth. In addition to $\mathcal{L}_{\mathrm{DiceCE}}$, the total loss FT_DCE includes also $\mathcal{L}_{\mathrm{FT}}$, which is defined by Eq. (5).

$$\mathcal{L}_{\mathrm{FT}} = \left(1 - \mathrm{T}(\alpha,\beta)\right)^{1/\gamma} \tag{5}$$

with, $\mathrm{T}(\alpha,\beta)$ expressed by Eq. (6) and $\gamma$ set empirically.

$$\mathrm{T}(\alpha,\beta) = \frac{\sum_{i=1}^{N} p_i g_i}{\sum_{i=1}^{N} p_i g_i + \alpha\sum_{i=1}^{N} p_i(1-g_i) + \beta\sum_{i=1}^{N}(1-p_i)g_i} \tag{6}$$

where $\alpha$ and $\beta$ control the penalisation effect for false-positives (FPs) and false-negatives (FNs), respectively and are set empirically through hyperparameter optimisation.

## 3. Experiments

MambaX-Net was compared with three single-time-point SOTA baseline models (nnU-Net-V2, SwinUNETR-V2, SegMamba) as well as our previous dual-scan model (DSM). Performance was measured using the Dice Similarity Coefficient (DSC) and the 95th percentile Hausdorff Distance (HD95), expressed in millimetres (mm).

### 3.1. Datasets

Experiments were performed on two datasets: the publicly available PI-CAI dataset (Saha et al., 2024), and the in-house (private) AS dataset, which was collected at Guy's and St Thomas' Hospital with the necessary consent and ethics approval for analysis granted by the Guy's Cancer Cohort ethics committee (REC 23/NW/0105). The PI-CAI dataset was utilised for pre-training and validation of the MambaX-Net backbone, while the AS dataset was employed for model fine-tuning and testing.

The PI-CAI dataset comprises 1500 bpMRI scans (from 1476 patients) acquired at either 1.5T or 3T, and we utilised the T2-weighted (T2W) images for our experiments. Within the dataset, expert-annotated labels for prostate zones are available for 195 patients originating from the earlier ProstateX dataset (total 328 patients) (Litjens et al., 2017). The remainder of the PI-CAI dataset contains only AI-generated segmentation masks. To prevent data leakage, we excluded all (328) ProstateX patient IDs from the PI-CAI dataset to obtain $PICAI_{train}$ containing the AI-labelled portion (1170 scans). $PICAI_{train}$ was used to train and optimise the nnU-Net backbone of MambaX-Net, using 5-fold cross-validation grouped by patient ID. The expert-labelled subset of 195 patients/scans, denoted as $PICAI_{test}$, was reserved as an external test set to verify that the backbone model achieved performance consistent with SOTA results reported in the literature.

The in-house AS dataset consists of longitudinal scans from 317 patients acquired during routine clinical practice over ten years (2012–2022) using Siemens 1.5T and 3T scanners, predominantly at 1.5T. Two board-certified urogenital radiologists annotated the WP, PZ and TZ for 92 patients, with one annotating 66 patients and the other annotating 26 patients. Manual segmentation required approximately 10 min per scan on average, with contouring of the apex and base being the most time-consuming step, highlighting the substantial clinical burden of this task. As this is a real-world dataset rather than a curated clinical trial cohort, the randomly selected subset of 92 patients reflects routine clinical MRI practice and includes a relatively high proportion of challenging cases. This includes patients who underwent treatments such as Transurethral Resection of the Prostate (TURP), and those with complex prostate boundary delineation and low signal-to-noise ratio (SNR). Among the 92 patients, 66 patients with three time points per patient, corresponding to 198 scans, were used as an independent test set ($AS_{test}$) for final model comparison and generalisation assessment. The remaining 26 patients with two time points per patient, corresponding to 52 scans ($AS_{val}$), were used for hyperparameter optimisation of all models. To reduce reader-specific bias, all annotated cases were first pooled and then randomly split into the validation and test sets. In addition to the 92 expert-annotated patients used for validation and testing, a separate set of 225 patients, corresponding to 450 scans and denoted as $AS_{train}$, had no expert annotation and was used to train MambaX-Net and the comparative baseline models using pseudo-labels generated by the pretrained nnU-Net backbone.

### 3.2. Pre-processing

All models used the same pre-processing pipeline: images were z-normalised, resampled to $0.5 \times 0.5 \times 3.0$ mm voxel spacing, axially cropped to $384 \times 384$, and processed as 3D patches with 16 slices. For dual-scan models, paired patches were extracted at matching z-coordinates to preserve spatial correspondence, while further alignment was performed in the latent feature space using the proposed M-CAM. SwinUNETR was the only exception, requiring $z$-axis padding to 32 slices to satisfy its minimum input depth.

### 3.3. Implementation details

The MambaX-Net experimental pipeline has two stages: First, we pre-trained a standard 3D nnU-Net on $PICAI_{train}$ (denoted as nnU-Net$_{\text{PICAI}}$), which serves two key purposes: (i) it provides the initial weights for the MambaX-Net encoder and decoder and (ii) produces pseudo-labels for $AS_{train}$ at two time points: $t = 0$ (denoted $M_0$) and $t = 1$ (denoted $M_1$). Then, using these pseudo-labels, we trained the M-CAM and SEM modules while fine-tuning the MambaX-Net encoder and decoder, which were initialised with nnU-Net$_{\text{PICAI}}$ weights. During training, $M_0$ is provided as input to the model alongside $I_0$ and $I_1$, while $M_1$ is used exclusively as ground truth for loss calculation, thereby guiding the semi-supervised learning process.

While relying on the nnU-Net-generated pseudo-labels might introduce a risk of systematic bias, deploying the fine-tuned nnU-Net backbone directly would yield suboptimal results, as seen by the zero-shot (n = 0) performance of nnU-Net in Tables 1–3. The generated pseudo-labels provide a scalable supervisory signal but remain imperfect, particularly at the apex/base and in the PZ, where they frequently yield anatomically inconsistent boundaries (Tables 1–4). MambaX-Net is designed to reduce the impact of such errors by exploiting patient-specific longitudinal context rather than relying on pseudo-labels alone. Specifically, the previous segmentation mask, although potentially imperfect, is encoded by SEM as a latent anatomical prior and fused with current-scan features through M-CAM. Thus, the final prediction remains driven by the current MRI while being regularised by temporally related anatomical context from the paired scan, a mechanism unavailable to the single-scan nnU-Net backbone. This architectural advantage is empirically supported by the zonal and regional results for n = 50 at $t = 1$ (Table 4), where, when both models are trained using the same pseudo-labels, MambaX-Net (n = 50) significantly outperforms the nnU-Net$_{\text{AS}}$ backbone fine-tuned with the same n = 50 pseudo-label cases and even exceeds the performance of nnU-Net$_{\text{AS}}$ fine-tuned with n = 225. These findings indicate that the pseudo-labels provide useful, scalable supervision, while the proposed dual-scan architecture contributes additional temporal and anatomical context beyond the backbone segmenter.

Notably, we did not perform image registration between $I_0$, $M_0$, and $I_1$ before feeding them to the model. The M-CAM using cross-attention is designed to implicitly align features in the latent space, as confirmed by the ablation test results.

To ensure a fair and rigorous comparison, all baseline models followed a similar two-stage training protocol described as follows:

- **nnU-Net, SwinUNETR, and SegMamba:** Each model was first pre-trained on the $PICAI_{train}$ dataset. The pre-trained nnU-Net$_{\text{PICAI}}$ model was empirically chosen to generate pseudo-labels for $AS_{train}$ based on the $PICAI_{test}$ results in Table 5, where nnU-Net$_{\text{PICAI}}$ significantly outperformed other comparative models. Since these are single-time-point models, the two longitudinal scans ($I_0$, $I_1$) were treated as independent scans rather than as paired sequential images.
- **DSM:** As a dual-scan model, DSM followed the same training protocol as MambaX-Net. Its encoder and decoder were initialised with nnU-Net$_{\text{PICAI}}$ weights, and paired scans $(I_0, I_1)$ from $AS_{train}$ were used for fine-tuning the encoder–decoder blocks and training the rest of the modules. Unlike MambaX-Net, DSM does not use $M_0$ as input and only uses $M_1$ for loss calculation.

We use the subscripts PICAI and AS to denote models trained on the $PICAI_{train}$ and $AS_{train}$ datasets, respectively (e.g., $model_name_{PICAI}$ and $model_name_{AS}$).

All models were trained using the stochastic gradient descent (SGD) optimiser alongside validation-based early stopping. Models trained on $PICAI_{train}$ ($model_name_{PICAI}$) ran for 1000 epochs with a polynomial decay learning-rate scheduler, while models trained on $AS_{train}$ ($model_name_{AS}$) were fine-tuned for 50 epochs using a OneCycle learning rate scheduler. OneCycle was chosen for the $AS_{train}$ dataset because it promotes super-convergence and improves performance when training data is limited (Smith and Topin, 2019). During training, spatial and intensity-based data augmentations were applied to prevent overfitting. Hyperparameter optimisation was performed using Optuna (Akiba et al., 2019) with $PICAI_{val}$ and $AS_{val}$ used for $model_name_{PICAI}$ and $model_name_{AS}$, respectively. To enhance the model's robustness and improve prediction accuracy during inference, we apply test-time augmentation (TTA) with random flips and rotations, and average the predictions to get the final result. Software versions, hardware details, and carbon-footprint reporting are provided in the Supplementary Material.

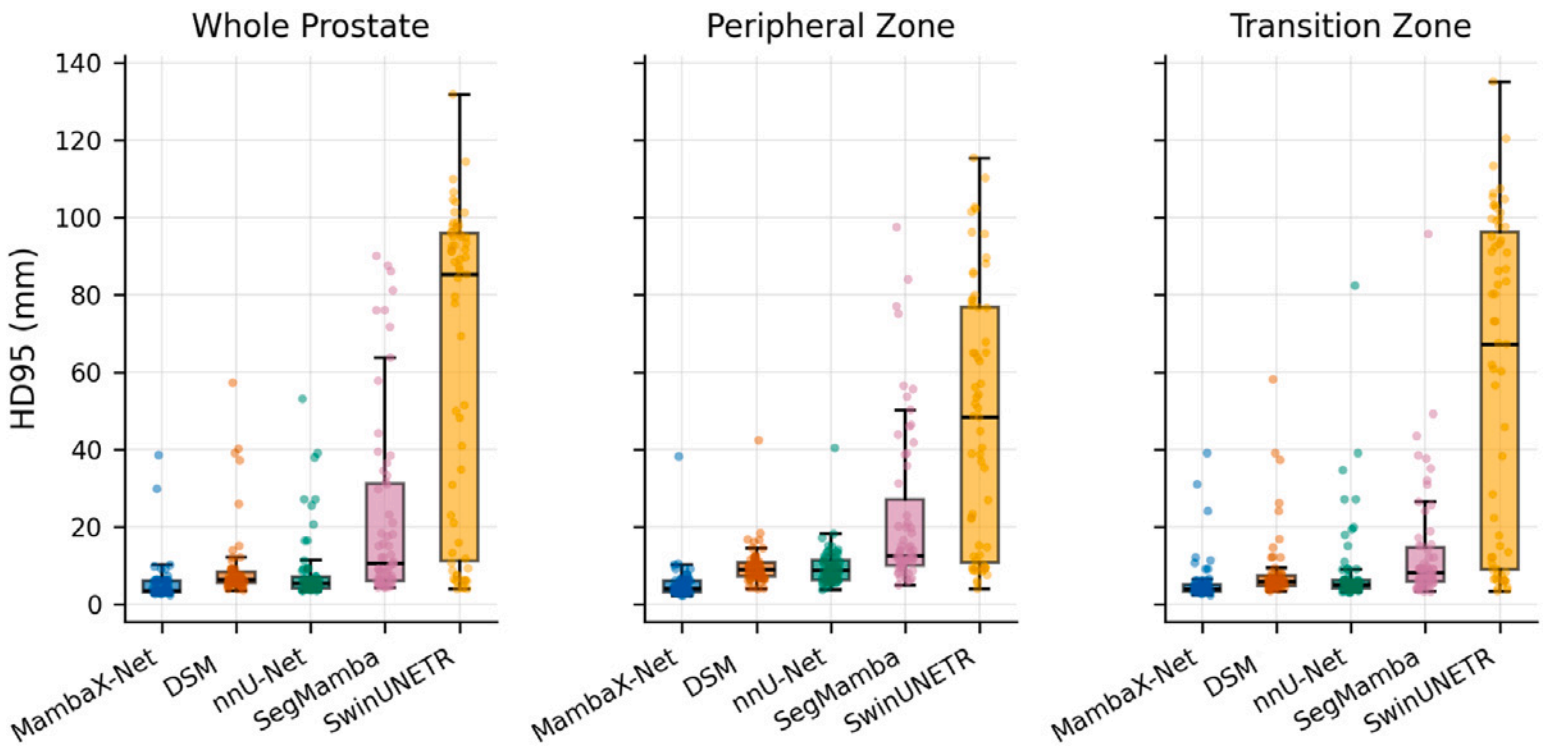

**Fig. 3.** Distribution of HD95 computed on $AS_{test}$ at $t = 1$ across WP, PZ, and TZ for the models trained on $AS_{train}$ with $n = 50$.

## 4. Results

In this section, we present the experimental results evaluating MambaX-Net, with performance measured using DSC and HD95 across the whole prostate (WP), peripheral zone (PZ), and transition zone (TZ). First, we compared MambaX-Net performance against SOTA models on $AS_{test}$ and analysed the impact of pseudo-labels quality on models' performance. Then, we validated the choice of nnU-Net as the optimal backbone for MambaX-Net and evaluated the contribution of individual architectural components by means of an ablation study. Results on $PICAI_{test}$ and $AS_{val}$ are reported as the mean ± standard deviation of the 3D metrics calculated across all patients, illustrating inter-patient performance variability. For the final held-out $AS_{test}$ set, results are reported as the mean with 95% confidence intervals in brackets, providing an estimate of the uncertainty around the mean performance and allowing the magnitude of differences between methods to be assessed. While the standard deviation highlights case-level heterogeneity, 95% confidence intervals were prioritised for the final test set to quantify the uncertainty bounds around the final mean performance estimate. To calculate these bounds without assuming a specific sampling distribution, all 95% confidence intervals were estimated via the percentile bootstrap using 10,000 resamples. The best-performing model for each metric and pseudo-label n group was determined by selecting the highest unrounded mean DSC or the lowest unrounded mean HD95 across all held-out test subjects. Finally, to assess statistical differences ($\alpha < 0.05$), the same statistical analysis was applied to all comparisons. Specifically, group-wise comparisons were performed using the Friedman rank-sum test followed by Holm's post-hoc analysis for multiple comparisons.

### *4.1. MambaX-Net comparison with SOTA models*

Table 1 summarises the performance at $t = 1$ of MambaX-Net and the comparative models fine-tuned with different numbers $n$ of patients from $AS_{train}$ ($n$ = 0, 5, 10, 50, 100 and 225). Here, $n = 0$ denotes the out-of-the-box performance of models pre-trained on $PICAI_{train}$ without any fine-tuning on $AS_{train}$. Results for the dual-scan models (DSM and MambaX-Net) are omitted at $n = 0$, since their additional components require training on paired scans from two time points, only available in $AS_{train}$.

For $n = 0$, the three single-scan models showed markedly different levels of robustness to domain shift. The performance of $\text{nnU-Net}_{\text{PICAI}}$, $\text{SwinUNETR}_{\text{PICAI}}$ and $\text{SegMamba}_{\text{PICAI}}$ declined significantly, with $\text{SegMamba}_{\text{PICAI}}$ showing the largest overall performance drop and the poorest PZ segmentation performance among the three models. Although fine-tuning on $AS_{train}$ ($n > 0$) improved the performance of all three single-scan models, none matched their original $PICAI_{test}$ performance, highlighting the challenge of domain adaptation and motivating strategies like dual-scan architectures.

For $n \leq 50$, MambaX-Net consistently outperformed all comparative models, achieving the highest DSC and lowest HD95 scores, especially in the PZ and TZ, with all differences being statistically significant. The next-best models were $\text{nnU-Net}_{\text{AS}}$ and DSM, which exhibited comparable performance among themselves but markedly lower accuracy than MambaX-Net. In contrast, $\text{SwinUNETR}_{\text{AS}}$ experienced a severe degradation in performance upon fine-tuning. Despite holding the most robust HD95 scores at zero-shot (n = 0), the introduction of $\text{nnU-Net}_{\text{PICAI}}$ generated noisy pseudo-labels, causing its DSC to drop below 0.35 across all zones, a failure to adapt that persisted regardless of training size.

As the number of patients used for fine-tuning grows ($n \geq 100$), the performance gap between most models diminishes for both DSC and HD95, except for $\text{SwinUNETR}_{\text{AS}}$, whose performance remains largely unchanged and consistently low across all training sizes. MambaX-Net always achieved either the best or second-best performance in DSC for larger $n$ values, demonstrating consistent competitiveness, albeit without statistically significant improvements. DSM's performance in DSC was roughly equal to $\text{nnU-Net}_{\text{AS}}$ and MambaX-Net in the WP, and ranked third in performance for PZ. These results demonstrate that the DSM's architectural advantage diminishes compared to SegMamba and nnU-Net when training on larger datasets. For HD95, MambaX-Net had the best performance in the PZ and TZ, and was outranked by DSM for the WP, but the difference was not statistically significant.

To further investigate the variability observed in HD95 across models and training set sizes, we analysed the distribution of HD95 values for each anatomical zone. The boxplots in Fig. 3 illustrate these distributions, with MambaX-Net achieving the lowest median and smallest interquartile range. This tightly clustered distribution indicates superior boundary accuracy and higher robustness across patients. By contrast, SegMamba and SwinUNETR exhibited substantially greater variability. Notably, SwinUNETR demonstrated the highest performance variability among all evaluated models, characterised by an exceptionally wide interquartile range and maximum HD95 values exceeding 115 mm in the PZ and 120 mm in both the WP and TZ. In the context of

**Table 1**
Segmentation performance of MambaX-Net and comparative models on $AS_{test}$ at $t = 1$. $n$ indicates the number of AS patients used for fine-tuning. Values are the metric mean (95% CI). For each metric, the best results are shown in **bold**.

| n | Model | DSC ↑ | | | HD95 (mm) ↓ | | |
|---|---|---|---|---|---|---|---|
| | | WP | PZ | TZ | WP | PZ | TZ |
| 0 | nnU-Net$_{PICAI}$ | **0.46 (0.41, 0.50)** | 0.28 (0.24, 0.32) | **0.50 (0.45, 0.55)** | 97.24 (95.22, 99.04)* | 98.05 (95.44, 100.36)* | 92.34 (88.20, 95.96)* |
| | SwinUNETR$_{PICAI}$ | 0.42 (0.33, 0.52) | **0.34 (0.25, 0.42)** | 0.44 (0.35, 0.53) | **47.35 (37.73, 56.99)** | **59.22 (48.12, 70.16)** | **40.62 (32.64, 48.77)** |
| | SegMamba$_{PICAI}$ | 0.14 (0.12, 0.15)* | 0.02 (0.02, 0.03)* | 0.16 (0.14, 0.18)* | 94.37 (92.90, 95.71)* | 92.46 (90.12, 94.68)* | 95.18 (93.66, 96.67)* |
| 5 | nnU-Net$_{AS}$ | 0.84 (0.82, 0.86) | 0.66 (0.62, 0.69)* | 0.77 (0.75, 0.79)* | 15.05 (10.25, 20.54) | 13.02 (9.92, 16.66)* | 16.09 (10.62, 22.42)* |
| | SwinUNETR$_{AS}$ | 0.25 (0.18, 0.33)* | 0.09 (0.06, 0.13)* | 0.20 (0.14, 0.28)* | 66.52 (59.81, 72.68)* | 56.89 (49.75, 63.76)* | 62.78 (53.97, 71.12)* |
| | SegMamba$_{AS}$ | 0.52 (0.47, 0.57)* | 0.23 (0.19, 0.26)* | 0.54 (0.49, 0.59)* | 83.79 (79.26, 87.98)* | 57.57 (51.00, 64.05)* | 50.00 (42.09, 58.25)* |
| | DSM | 0.80 (0.78, 0.82)* | 0.59 (0.56, 0.62)* | 0.78 (0.76, 0.80)* | 10.34 (7.89, 13.21)* | 9.20 (7.50, 11.34)* | 10.89 (7.97, 14.55)* |
| | MambaX-Net | **0.84 (0.82, 0.86)** | **0.68 (0.65, 0.71)** | **0.82 (0.80, 0.84)** | **8.06 (6.10, 10.45)** | **7.71 (6.23, 9.50)** | **9.34 (6.08, 13.56)** |
| 10 | nnU-Net$_{AS}$ | 0.82 (0.79, 0.83)* | 0.57 (0.52, 0.61)* | 0.79 (0.77, 0.81)* | 12.33 (8.69, 16.66)* | 10.83 (8.49, 13.73)* | 11.00 (7.35, 15.45)* |
| | SwinUNETR$_{AS}$ | 0.27 (0.19, 0.35)* | 0.11 (0.07, 0.15)* | 0.22 (0.15, 0.30)* | 58.06 (50.98, 64.84)* | 50.26 (43.55, 56.88)* | 55.62 (47.21, 63.81)* |
| | SegMamba$_{AS}$ | 0.64 (0.60, 0.68)* | 0.29 (0.25, 0.33)* | 0.63 (0.58, 0.66)* | 54.38 (46.46, 62.29)* | 38.67 (31.85, 45.75)* | 26.53 (20.66, 33.30)* |
| | DSM | 0.83 (0.82, 0.85)* | 0.65 (0.62, 0.68)* | 0.80 (0.78, 0.82)* | 9.02 (6.79, 11.73)* | 6.94 (5.88, 8.42)* | 9.23 (7.03, 11.91)* |
| | MambaX-Net | **0.86 (0.84, 0.88)** | **0.73 (0.70, 0.75)** | **0.82 (0.80, 0.84)** | **8.01 (5.87, 10.63)** | **6.22 (5.13, 7.61)** | **7.99 (5.79, 10.76)** |
| 50 | nnU-Net$_{AS}$ | 0.84 (0.83, 0.85)* | 0.58 (0.55, 0.61)* | 0.81 (0.80, 0.83)* | 8.77 (6.58, 11.32)* | 9.52 (8.41, 10.91)* | 8.87 (6.33, 12.16)* |
| | SwinUNETR$_{AS}$ | 0.33 (0.24, 0.43)* | 0.19 (0.13, 0.25)* | 0.32 (0.23, 0.41)* | 61.34 (50.93, 71.40)* | 46.87 (38.67, 55.09)* | 56.99 (46.59, 67.33)* |
| | SegMamba$_{AS}$ | 0.76 (0.73, 0.78)* | 0.40 (0.37, 0.44)* | 0.73 (0.70, 0.76)* | 22.46 (16.70, 29.01)* | 22.68 (17.78, 28.30)* | 13.40 (10.13, 17.56)* |
| | DSM | 0.80 (0.78, 0.81)* | 0.50 (0.47, 0.52)* | 0.79 (0.77, 0.81)* | 9.39 (7.25, 12.03)* | 9.76 (8.67, 11.19)* | 8.64 (6.58, 11.21)* |
| | MambaX-Net | **0.90 (0.89, 0.90)** | **0.77 (0.75, 0.79)** | **0.87 (0.86, 0.88)** | **5.20 (4.03, 6.75)** | **5.14 (4.24, 6.47)** | **5.80 (4.45, 7.51)** |
| 100 | nnU-Net$_{AS}$ | **0.83 (0.82, 0.85)** | 0.63 (0.61, 0.66)* | 0.81 (0.79, 0.83)* | 8.75 (6.69, 11.16) | 8.64 (7.35, 10.23)* | 10.47 (7.09, 14.54) |
| | SwinUNETR$_{AS}$ | 0.34 (0.24, 0.43)* | 0.18 (0.13, 0.24)* | 0.32 (0.22, 0.41)* | 61.66 (50.86, 72.19)* | 45.90 (37.69, 54.46)* | 55.20 (44.61, 65.78)* |
| | SegMamba$_{AS}$ | 0.79 (0.76, 0.81)* | 0.44 (0.41, 0.47)* | 0.76 (0.73, 0.78)* | 15.28 (11.37, 19.90)* | 14.70 (12.26, 17.68)* | 12.06 (9.07, 15.88)* |
| | DSM | 0.83 (0.82, 0.84) | 0.50 (0.48, 0.53)* | 0.81 (0.80, 0.83) | **8.48 (6.59, 10.75)** | 9.68 (8.56, 11.10)* | 8.12 (6.33, 10.28) |
| | MambaX-Net | 0.83 (0.81, 0.85) | **0.66 (0.62, 0.69)** | **0.82 (0.80, 0.84)** | 8.51 (6.64, 10.76) | **7.49 (6.20, 9.10)** | **8.07 (5.64, 11.44)** |
| 238 | nnU-Net$_{AS}$ | 0.83 (0.81, 0.84) | **0.65 (0.62, 0.68)** | 0.80 (0.78, 0.81)* | 11.61 (8.13, 15.63) | 9.34 (7.23, 12.23)* | 13.45 (8.54, 19.38)* |
| | SwinUNETR$_{AS}$ | 0.34 (0.24, 0.43)* | 0.16 (0.11, 0.22)* | 0.31 (0.22, 0.41)* | 57.22 (47.39, 66.59)* | 46.13 (38.63, 53.77)* | 56.95 (46.54, 67.04)* |
| | SegMamba$_{AS}$ | 0.78 (0.75, 0.81)* | 0.47 (0.43, 0.50)* | 0.75 (0.72, 0.78)* | 14.20 (10.49, 18.69)* | 12.64 (10.55, 15.06)* | 9.09 (7.49, 10.91)* |
| | DSM | 0.82 (0.81, 0.83)* | 0.50 (0.48, 0.53)* | 0.81 (0.80, 0.82)* | **7.13 (5.90, 8.69)** | 9.61 (8.51, 11.02)* | 7.11 (5.72, 8.81)* |
| | MambaX-Net | **0.83 (0.82, 0.84)** | 0.59 (0.56, 0.62)* | **0.84 (0.82, 0.85)** | 7.48 (5.89, 9.45) | **9.04 (7.92, 10.44)** | **6.23 (4.90, 7.95)** |

* Indicates a statistically significant difference ($\alpha < 0.05$) from the best result within the same $n$ group.

**Table 2**
Segmentation performance of MambaX-Net and comparative models on $AS_{test}$ at $t = 2$. $n$ indicates the number of AS patients used for fine-tuning. Values are mean (95% CI). For each metric, the best results are shown in **bold**.

| n | Model | DSC ↑ | | | HD95 (mm) ↓ | | |
|---|---|---|---|---|---|---|---|
| | | WP | PZ | TZ | WP | PZ | TZ |
| 0 | nnU-Net$_{PICAI}$ | **0.46 (0.42, 0.50)** | 0.26 (0.22, 0.30) | **0.50 (0.46, 0.55)** | 96.36 (93.75, 98.79)* | 98.92 (95.90, 101.82)* | 89.25 (84.60, 93.39)* |
| | SwinUNETR$_{PICAI}$ | 0.40 (0.30, 0.50) | **0.33 (0.24, 0.43)** | 0.41 (0.31, 0.50)* | **50.50 (40.17, 60.86)** | **60.12 (48.44, 71.41)** | **44.46 (35.52, 53.29)** |
| | SegMamba$_{PICAI}$ | 0.15 (0.13, 0.17)* | 0.02 (0.01, 0.03)* | 0.17 (0.15, 0.20)* | 93.39 (91.69, 94.99)* | 90.44 (88.06, 92.79)* | 94.17 (92.48, 95.78)* |
| 5 | nnU-Net$_{AS}$ | **0.85 (0.83, 0.86)** | **0.65 (0.61, 0.68)** | 0.78 (0.76, 0.80)* | 15.07 (10.72, 20.04) | 10.70 (8.89, 12.71)* | 13.97 (10.06, 18.43)* |
| | SwinUNETR$_{AS}$ | 0.28 (0.20, 0.36)* | 0.11 (0.07, 0.15)* | 0.23 (0.16, 0.31)* | 67.35 (61.18, 73.14)* | 56.86 (49.86, 63.55)* | 59.44 (50.80, 67.76)* |
| | SegMamba$_{AS}$ | 0.53 (0.49, 0.58)* | 0.21 (0.18, 0.24)* | 0.56 (0.51, 0.60)* | 82.68 (78.07, 86.90)* | 60.94 (55.27, 66.29)* | 52.05 (45.07, 58.97)* |
| | DSM | 0.81 (0.79, 0.82)* | 0.58 (0.55, 0.61)* | 0.79 (0.76, 0.81)* | **8.46 (6.94, 10.30)** | 9.20 (7.53, 11.27) | **8.84 (6.88, 11.22)** |
| | MambaX-Net | 0.83 (0.80, 0.85) | 0.65 (0.60, 0.69) | **0.81 (0.79, 0.84)** | 8.76 (6.47, 11.55)* | **8.32 (6.86, 10.08)** | 13.16 (8.51, 18.50) |
| 10 | nnU-Net$_{AS}$ | 0.82 (0.80, 0.84)* | 0.56 (0.52, 0.59)* | 0.80 (0.78, 0.82)* | 10.06 (7.44, 13.14)* | 9.29 (7.92, 10.91)* | 9.13 (6.83, 11.87)* |
| | SwinUNETR$_{AS}$ | 0.29 (0.21, 0.37)* | 0.12 (0.08, 0.17)* | 0.24 (0.16, 0.32)* | 58.93 (51.59, 65.88)* | 49.99 (43.01, 56.67)* | 54.41 (46.11, 62.47)* |
| | SegMamba$_{AS}$ | 0.66 (0.62, 0.70)* | 0.28 (0.24, 0.32)* | 0.65 (0.61, 0.68)* | 55.18 (47.30, 62.70)* | 37.79 (31.66, 44.16)* | 25.96 (20.08, 32.32)* |
| | DSM | 0.84 (0.82, 0.85)* | 0.64 (0.61, 0.67)* | 0.82 (0.80, 0.83)* | 7.64 (6.01, 9.64)* | 8.18 (6.46, 10.35)* | 7.02 (5.71, 8.72) |
| | MambaX-Net | **0.86 (0.84, 0.88)** | **0.71 (0.68, 0.74)** | **0.83 (0.82, 0.85)** | **6.37 (5.14, 7.98)** | **6.86 (5.54, 8.50)** | **6.60 (5.35, 8.21)** |
| 50 | nnU-Net$_{AS}$ | 0.84 (0.83, 0.85)* | 0.57 (0.54, 0.60)* | 0.82 (0.80, 0.83)* | 8.42 (6.47, 10.85)* | 10.88 (8.74, 13.73)* | 8.21 (6.24, 10.56)* |
| | SwinUNETR$_{AS}$ | 0.33 (0.24, 0.43)* | 0.21 (0.15, 0.28)* | 0.31 (0.22, 0.41)* | 60.66 (49.77, 71.27)* | 45.71 (37.52, 53.78)* | 55.53 (45.08, 65.73)* |
| | SegMamba$_{AS}$ | 0.77 (0.74, 0.80)* | 0.40 (0.36, 0.43)* | 0.74 (0.71, 0.77)* | 18.75 (14.34, 23.69)* | 19.53 (15.40, 24.30)* | 12.48 (9.59, 15.94)* |
| | DSM | 0.81 (0.79, 0.82)* | 0.50 (0.47, 0.52)* | 0.81 (0.79, 0.82)* | 7.83 (6.41, 9.66)* | 10.51 (9.25, 12.08)* | 7.63 (5.97, 9.67)* |
| | MambaX-Net | **0.90 (0.89, 0.91)** | **0.77 (0.74, 0.79)** | **0.87 (0.86, 0.88)** | **4.41 (3.79, 5.20)** | **4.92 (4.35, 5.52)** | **5.60 (4.26, 7.35)** |
| 100 | nnU-Net$_{AS}$ | 0.83 (0.82, 0.85) | 0.62 (0.59, 0.65)* | 0.82 (0.80, 0.83)* | 8.71 (6.80, 10.92)* | 9.44 (7.59, 11.88)* | 9.24 (6.95, 11.84)* |
| | SwinUNETR$_{AS}$ | 0.34 (0.24, 0.44)* | 0.19 (0.13, 0.26)* | 0.31 (0.22, 0.41)* | 61.50 (50.16, 72.40)* | 44.81 (36.42, 53.31)* | 54.86 (44.52, 65.18)* |
| | SegMamba$_{AS}$ | 0.79 (0.77, 0.81)* | 0.43 (0.39, 0.46)* | 0.76 (0.74, 0.79)* | 14.86 (11.15, 19.04)* | 16.60 (13.18, 20.63)* | 11.02 (8.26, 14.37)* |
| | DSM | **0.83 (0.82, 0.85)** | 0.51 (0.48, 0.54)* | 0.82 (0.81, 0.83) | **6.03 (5.33, 6.83)** | 9.60 (8.73, 10.51)* | **6.64 (5.30, 8.40)** |
| | MambaX-Net | 0.83 (0.81, 0.85) | **0.64 (0.60, 0.67)** | **0.83 (0.81, 0.84)** | 7.72 (6.19, 9.63)* | **8.23 (6.88, 9.92)** | 7.35 (5.71, 9.30) |
| 238 | nnU-Net$_{AS}$ | 0.83 (0.81, 0.84) | **0.63 (0.60, 0.66)** | 0.80 (0.78, 0.82)* | 9.90 (7.44, 12.84) | 8.96 (7.12, 11.41)* | 11.21 (8.08, 14.91)* |
| | SwinUNETR$_{AS}$ | 0.36 (0.26, 0.46)* | 0.19 (0.13, 0.25)* | 0.33 (0.24, 0.43)* | 55.57 (45.28, 65.38)* | 43.20 (35.43, 51.10)* | 54.12 (43.58, 64.41)* |
| | SegMamba$_{AS}$ | 0.81 (0.79, 0.82)* | 0.47 (0.44, 0.50)* | 0.77 (0.75, 0.79)* | 13.61 (10.11, 17.61)* | 12.65 (10.24, 15.59)* | 10.04 (7.97, 12.48)* |
| | DSM | 0.83 (0.81, 0.84) | 0.51 (0.48, 0.54)* | 0.82 (0.80, 0.83)* | 6.15 (5.47, 6.96) | 9.42 (8.59, 10.28)* | 6.73 (5.36, 8.51)* |
| | MambaX-Net | **0.83 (0.82, 0.84)** | 0.58 (0.55, 0.61)* | **0.84 (0.82, 0.85)** | **6.03 (5.35, 6.83)** | **8.81 (7.98, 9.69)** | **6.27 (4.86, 8.11)** |

* Indicates a statistically significant difference ($\alpha < 0.05$) from the best result within the same $n$ group.

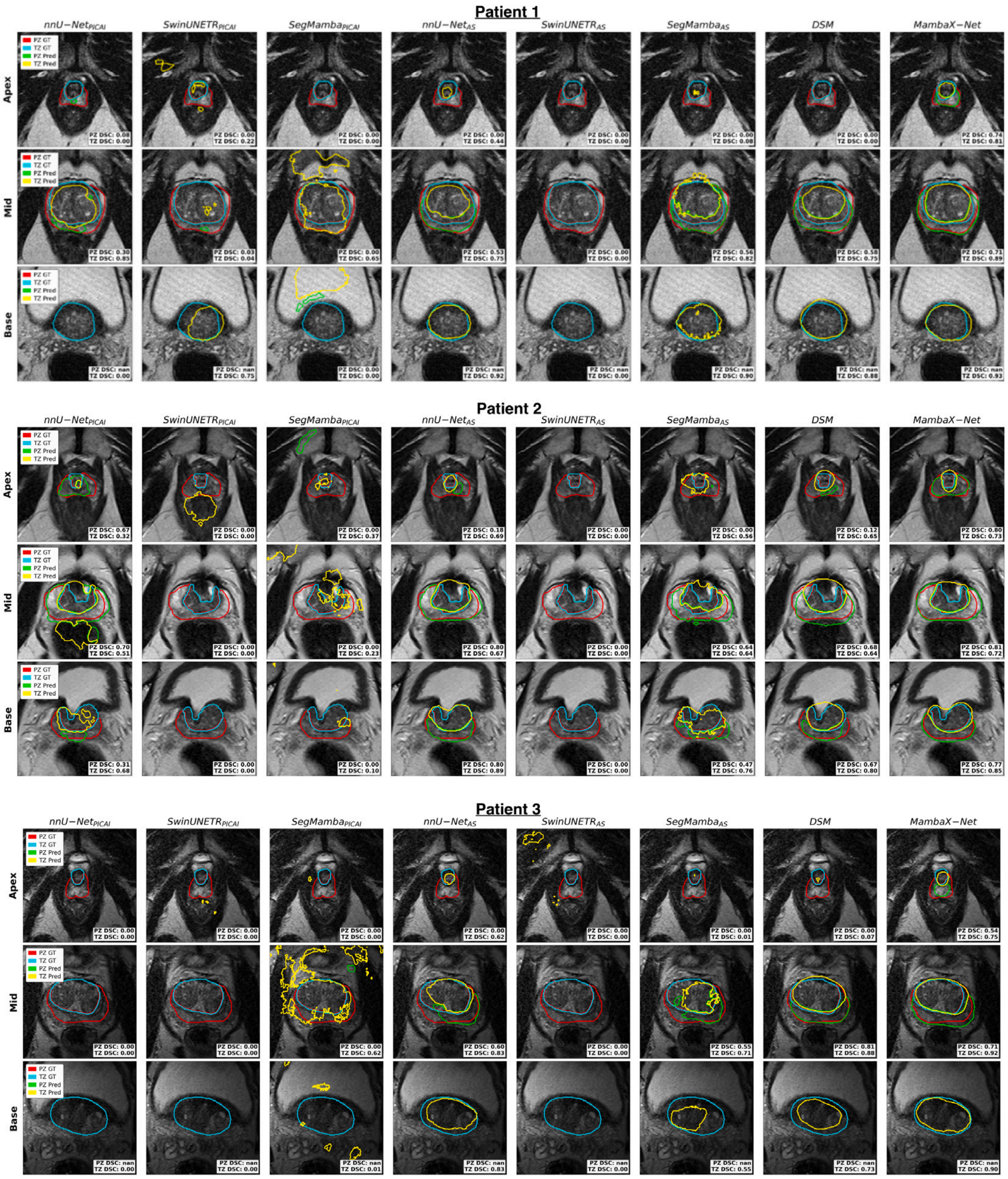

**Fig. 4.** Qualitative comparison of all models ($n = 50$) on three $AS_{test}$ patients at t = 1.

prostate MRI, boundary distance errors of this magnitude suggest that the model is not merely struggling with precise boundary delineation, but is frequently hallucinating and predicting foreground labels on entirely irrelevant anatomical structures.

We also tested the MambaX-Net fine-tuned on $I_0$ and $I_1$ directly on further time points $t = 2$ and $t = 3$ of the 66 patients in $AS_{test}$ without further fine-tuning, with the results displayed in Tables 2–3. Similar to Table 1, at $n = 0$ in Tables 2–3, the three single-scan models showed different levels of robustness, with SwinUNETR$_{\text{PICAI}}$ and nnU-Net$_{\text{PICAI}}$ showing higher performance compared to SegMamba$_{\text{PICAI}}$.

For $n \le 50$ in Tables 2–3, MambaX-Net consistently outperformed all comparative models and achieved the highest DSC scores, except for $n = 5$ at $t = 2$, where nnU-Net$_{\text{AS}}$ had equal performance with no statistically significant difference. Boundary distance performance measured by HD95 showed that MambaX-Net outperformed comparative models, only being outperformed by DSM for $n = 5$ in the WP and TZ at $t = 2$, and for $n = 10$ for PZ at $t = 3$. SwinUNETR$_{\text{AS}}$ was the worst-performing model for all metrics. Similar to Table 1, DSM's performance in DSC was roughly equal to nnU-Net$_{\text{AS}}$ in the WP and TZ, and ranked third in performance for PZ, with the performance advantage diminishing after $n > 10$. For $n \ge 100$, MambaX-Net still retained the best or second-best DSC and HD95 performance.

Fig. 4 shows a qualitative comparison of the prostate segmentations for three test patients with ground-truth (GT) and model predictions for the WP, PZ, and TZ overlaid on axial T2-weighted images. Specifically, Patient 1 represents a typical AS scan with standard anatomy but moderate noise. Patient 2 illustrates performance on post-surgical anatomy; the patient has undergone a TURP procedure, which is visible at the prostate base and mid-gland. Patient 3 has a hip artefact, causing uneven (gradient) image intensity and poor contrast within the prostate. As shown in Fig. 4, baseline models trained exclusively on PI-CAI exhibit severe boundary leakage when tested on $AS_{test}$. Even after fine-tuning, SwinUNETR$_{\text{AS}}$ and SegMamba$_{\text{AS}}$ exhibit disjointed predictions and fail to delineate complex boundaries, particularly under the gradient artefact of Patient 3. Following fine-tuning on the AS dataset, nnU-Net$_{\text{AS}}$ adapts to the surgical defect in Patient 2, but

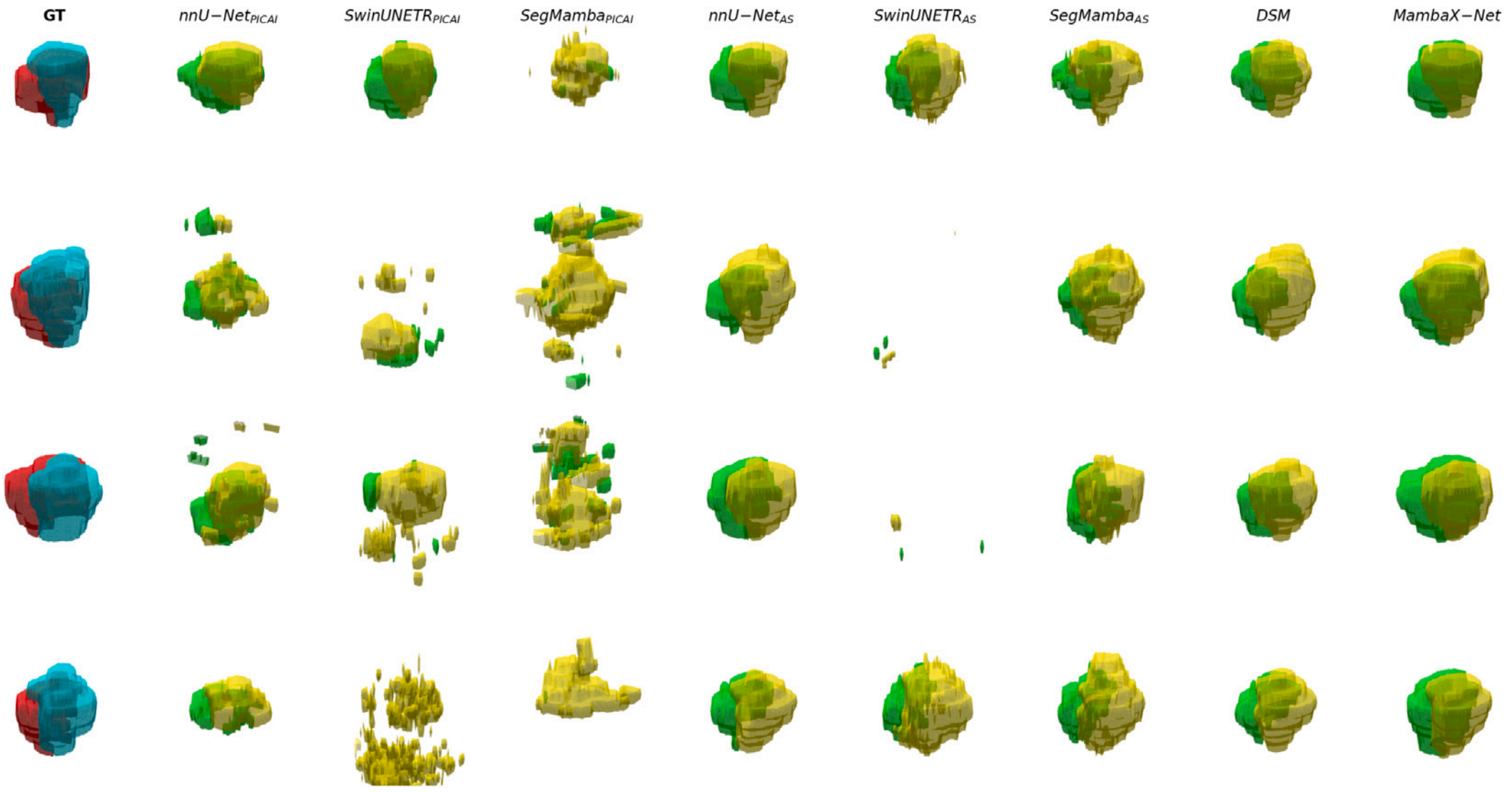

**Fig. 5.** Qualitative comparison of all models ($n = 50$) on four $AS_{test}$ patients at t = 1, with segmentations rendered as 3D volumes.

**Table 3**
Segmentation performance of MambaX-Net and comparative models on $AS_{test}$ at $t = 3$. $n$ Indicates the number of AS patients used for fine-tuning. Values are mean (95% CI). For each metric, the best results are shown in **bold**.

| | | DSC ↑ | | | HD95 (mm) ↓ | | |
|---|---|---|---|---|---|---|---|
| n | Model | WP | PZ | TZ | WP | PZ | TZ |
| 0 | nnU-Net$_{PICAI}$ | **0.46 (0.41, 0.50)** | 0.26 (0.22, 0.30) | **0.51 (0.47, 0.55)** | 97.82 (95.82, 99.74)* | 99.94 (97.52, 102.30)* | 93.13 (90.04, 96.15)* |
| | SwinUNETR$_{PICAI}$ | 0.43 (0.34, 0.53) | **0.36 (0.27, 0.44)** | 0.44 (0.35, 0.53) | **49.27 (39.56, 59.03)** | **58.90 (47.83, 69.82)** | **43.37 (35.04, 51.50)** |
| | SegMamba$_{PICAI}$ | 0.13 (0.11, 0.14)* | 0.02 (0.01, 0.02)* | 0.14 (0.12, 0.16)* | 94.40 (92.91, 95.83)* | 92.20 (90.10, 94.22)* | 95.27 (93.77, 96.77)* |
| 5 | nnU-Net$_{AS}$ | 0.82 (0.76, 0.86) | 0.62 (0.55, 0.68) | 0.77 (0.73, 0.80)* | 18.74 (8.86, 30.65) | 16.14 (8.72, 25.38)* | 16.26 (6.50, 29.22)* |
| | SwinUNETR$_{AS}$ | 0.35 (0.21, 0.49)* | 0.14 (0.07, 0.22)* | 0.30 (0.16, 0.44)* | 57.63 (44.07, 70.99)* | 53.64 (40.56, 67.35)* | 49.41 (34.39, 64.78)* |
| | SegMamba$_{AS}$ | 0.46 (0.39, 0.54)* | 0.20 (0.13, 0.27)* | 0.49 (0.42, 0.56)* | 81.53 (68.84, 91.93)* | 69.34 (59.18, 78.87)* | 59.46 (47.16, 72.05)* |
| | DSM | 0.78 (0.74, 0.81)* | 0.54 (0.49, 0.59)* | 0.75 (0.71, 0.78)* | 13.00 (7.88, 19.65)* | 13.01 (7.38, 20.76)* | 11.85 (7.43, 18.21)* |
| | MambaX-Net | **0.83 (0.80, 0.86)** | **0.63 (0.57, 0.69)** | **0.82 (0.79, 0.84)** | **9.74 (5.67, 15.35)** | **9.79 (5.67, 15.64)** | **8.16 (4.71, 14.12)** |
| 10 | nnU-Net$_{AS}$ | 0.80 (0.75, 0.84)* | 0.53 (0.46, 0.61)* | 0.78 (0.74, 0.81)* | 14.74 (6.32, 25.75)* | 14.14 (7.49, 23.95)* | 15.39 (6.04, 27.80)* |
| | SwinUNETR$_{AS}$ | 0.39 (0.25, 0.53)* | 0.15 (0.07, 0.23)* | 0.32 (0.18, 0.46)* | 52.75 (38.66, 66.28)* | 46.30 (34.56, 58.55)* | 42.05 (27.97, 56.75)* |
| | SegMamba$_{AS}$ | 0.61 (0.55, 0.66)* | 0.26 (0.18, 0.35)* | 0.61 (0.54, 0.66)* | 72.97 (58.37, 85.98)* | 57.04 (44.63, 68.37)* | 36.14 (23.31, 50.17)* |
| | DSM | 0.82 (0.78, 0.85)* | 0.61 (0.56, 0.66)* | 0.78 (0.73, 0.82)* | 8.52 (5.80, 12.16) | **8.33 (5.62, 12.70)** | 8.95 (6.00, 12.93)* |
| | MambaX-Net | **0.85 (0.82, 0.88)** | **0.69 (0.63, 0.74)** | **0.82 (0.79, 0.85)** | **7.78 (4.99, 11.78)** | 8.53 (4.83, 13.84) | **6.06 (4.51, 8.10)** |
| 50 | nnU-Net$_{AS}$ | 0.83 (0.79, 0.86)* | 0.55 (0.48, 0.62)* | 0.80 (0.77, 0.83)* | 13.24 (5.58, 23.72)* | 10.32 (7.44, 14.85)* | 7.14 (4.97, 10.23)* |
| | SwinUNETR$_{AS}$ | 0.44 (0.27, 0.59)* | 0.25 (0.15, 0.35)* | 0.40 (0.24, 0.55)* | 56.00 (38.24, 73.38)* | 44.09 (29.14, 59.51)* | 42.25 (25.48, 60.07)* |
| | SegMamba$_{AS}$ | 0.73 (0.68, 0.78)* | 0.37 (0.29, 0.44)* | 0.69 (0.64, 0.74)* | 24.78 (15.90, 34.35)* | 28.14 (17.80, 39.71)* | 16.91 (7.83, 28.01)* |
| | DSM | 0.80 (0.77, 0.82)* | 0.47 (0.42, 0.53)* | 0.78 (0.75, 0.80)* | 8.43 (5.94, 12.08)* | 12.68 (8.37, 18.41)* | 6.81 (5.65, 8.45)* |
| | MambaX-Net | **0.89 (0.87, 0.91)** | **0.75 (0.71, 0.79)** | **0.85 (0.83, 0.88)** | **7.41 (3.70, 12.49)** | **6.03 (3.84, 9.78)** | **6.50 (3.98, 10.91)** |
| 100 | nnU-Net$_{AS}$ | 0.82 (0.78, 0.85) | 0.59 (0.52, 0.65) | 0.80 (0.77, 0.82)* | 13.33 (6.06, 22.71) | 10.35 (6.71, 15.37) | 14.09 (5.83, 24.78) |
| | SwinUNETR$_{AS}$ | 0.44 (0.27, 0.59)* | 0.23 (0.13, 0.33)* | 0.41 (0.25, 0.56)* | 56.05 (36.57, 75.07)* | 43.44 (28.79, 58.86)* | 40.62 (23.78, 59.15)* |
| | SegMamba$_{AS}$ | 0.76 (0.70, 0.80)* | 0.39 (0.31, 0.47)* | 0.73 (0.67, 0.77)* | 17.60 (10.38, 25.78)* | 25.47 (15.73, 37.53)* | 15.07 (7.02, 25.79)* |
| | DSM | **0.82 (0.80, 0.85)** | 0.48 (0.42, 0.54)* | 0.79 (0.76, 0.82)* | **7.31 (5.25, 10.16)** | 11.25 (8.26, 15.34)* | 7.70 (5.51, 10.60) |
| | MambaX-Net | 0.82 (0.79, 0.85) | **0.61 (0.55, 0.67)** | **0.82 (0.79, 0.84)** | 10.63 (6.56, 15.66)* | **9.54 (6.13, 14.16)** | **6.01 (4.84, 7.46)** |
| 238 | nnU-Net$_{AS}$ | 0.80 (0.76, 0.84) | **0.60 (0.53, 0.66)** | 0.78 (0.75, 0.81)* | 16.10 (8.05, 26.12) | 12.02 (7.59, 17.83) | 13.79 (6.24, 24.42) |
| | SwinUNETR$_{AS}$ | 0.45 (0.29, 0.62)* | 0.22 (0.13, 0.32)* | 0.43 (0.26, 0.58)* | 49.41 (31.82, 67.05)* | 36.04 (24.62, 48.28)* | 43.64 (26.58, 61.47)* |
| | SegMamba$_{AS}$ | 0.77 (0.71, 0.81) | 0.43 (0.36, 0.50)* | 0.73 (0.68, 0.77)* | 16.32 (9.08, 24.73) | 20.84 (11.77, 31.91)* | 15.23 (7.94, 25.47)* |
| | DSM | 0.82 (0.79, 0.85) | 0.49 (0.43, 0.55)* | 0.80 (0.77, 0.82)* | **7.74 (5.36, 11.19)** | 10.80 (7.87, 14.90) | **6.14 (5.04, 7.63)** |
| | MambaX-Net | **0.82 (0.80, 0.85)** | 0.56 (0.51, 0.62)* | **0.82 (0.80, 0.84)** | 9.26 (5.34, 14.76) | **10.03 (7.15, 14.13)** | 7.40 (4.72, 12.05) |

* Indicates a statistically significant difference ($\alpha < 0.05$) from the best result within the same $n$ group.

under-segments at the apex of the prostate. Meanwhile, SwinUNETR$_{AS}$ and SegMamba$_{AS}$ continue to exhibit disjointed predictions and fail to delineate complex boundaries. MambaX-Net proves to be the most robust architecture across all scenarios, maintaining superior boundary delineation across all anatomical zones.

To further illustrate differences in model performance, Fig. 5 visualises the 3D segmentation volumes. These 3D renderings reveal that the pre-trained single-scan models ($model_{PICAI}$) frequently produce highly fragmented, anatomically implausible shapes, especially SwinUNETR$_{PICAI}$ and SegMamba$_{PICAI}$. However, SegMamba exhibits some degree of domain adaptation following fine-tuning, unlike SwinUNETR. Conversely, MambaX-Net generates smooth, contiguous volumes that accurately replicate the spatial geometry of the GT. This topological consistency visually confirms that the explicit anatomical prior injected by the SEM, combined with the M-CAM, effectively grounds the network and prevents severe volumetric hallucination.

A notable finding from these experiments is MambaX-Net's data efficiency. When trained with only $n = 10$ or $n = 50$, MambaX-Net surpasses all other models trained with significantly more data, $n \geq 100$. This advantage is clear across all quantitative metrics at all time points

**Table 4**

Regional and prostate volume stratified zone performance for $n = 50$ at $t = 1$. Panel A summarises regional (apex, mid-gland, base) DSC and HD95 with 95% confidence intervals for WP, TZ, and PZ. Panel B reports DSC and HD95 with 95% confidence intervals across zone-specific ground-truth volume tertiles for WP, TZ, and PZ. For each metric, the best results are shown in **bold**.

Panel A. Regional performance by anatomical zone

| Zone | Model | DSC ↑ | | | HD95 (mm) ↓ | | |
|---|---|---|---|---|---|---|---|
| | | Apex | Mid | Base | Apex | Mid | Base |
| WP | nnU-Net | 0.74 (0.71, 0.77)* | 0.90 (0.88, 0.90)* | 0.85 (0.83, 0.86)* | 3.81 (3.37, 4.31)* | 1.77 (1.41, 2.21)* | 2.68 (2.30, 3.12)* |
| | SwinUNETR | 0.31 (0.22, 0.40)* | 0.38 (0.28, 0.48)* | 0.34 (0.25, 0.44)* | 23.35 (18.67, 28.36)* | 34.77 (26.56, 43.11)* | 42.88 (33.33, 52.80)* |
| | SegMamba | 0.73 (0.69, 0.76)* | 0.84 (0.82, 0.86)* | 0.74 (0.70, 0.77)* | 5.66 (3.68, 9.12)* | 4.63 (2.76, 7.73)* | 5.46 (4.27, 6.92)* |
| | DSM | 0.68 (0.64, 0.71)* | 0.86 (0.85, 0.87)* | 0.79 (0.77, 0.81)* | 4.68 (4.23, 5.17)* | 2.73 (2.41, 3.09)* | 3.93 (3.31, 4.66)* |
| | MambaX-Net | **0.87 (0.86, 0.89)** | **0.94 (0.93, 0.94)** | **0.86 (0.85, 0.88)** | **1.82 (1.52, 2.16)** | **0.54 (0.37, 0.75)** | **2.25 (1.91, 2.63)** |
| TZ | nnU-Net | 0.74 (0.72, 0.77)* | 0.86 (0.85, 0.88)* | 0.83 (0.81, 0.84)* | 3.50 (3.07, 4.00)* | 2.09 (1.72, 2.59)* | 2.83 (2.44, 3.32)* |
| | SwinUNETR | 0.27 (0.19, 0.36)* | 0.35 (0.25, 0.45)* | 0.33 (0.24, 0.43)* | 28.17 (21.58, 35.21)* | 33.46 (26.03, 41.38)* | 34.58 (25.89, 43.78)* |
| | SegMamba | 0.65 (0.61, 0.69)* | 0.80 (0.78, 0.83)* | 0.73 (0.69, 0.77)* | 6.78 (4.52, 10.49)* | 4.67 (2.93, 7.81)* | 4.90 (4.05, 5.89)* |
| | DSM | 0.72 (0.69, 0.75)* | 0.85 (0.83, 0.86)* | 0.79 (0.77, 0.81)* | 3.71 (3.32, 4.15)* | 2.49 (2.16, 2.87)* | 3.66 (3.06, 4.47)* |
| | MambaX-Net | **0.82 (0.80, 0.83)** | **0.91 (0.90, 0.92)** | **0.86 (0.84, 0.87)** | **2.50 (2.19, 2.84)** | **1.06 (0.83, 1.31)** | **2.31 (2.01, 2.65)** |
| PZ | nnU-Net | 0.45 (0.40, 0.49)* | 0.66 (0.63, 0.69)* | 0.44 (0.38, 0.50)* | 8.43 (7.35, 9.59)* | 4.64 (4.03, 5.31)* | 6.88 (5.69, 8.17)* |
| | SwinUNETR | 0.12 (0.08, 0.17)* | 0.22 (0.15, 0.29)* | 0.15 (0.10, 0.21)* | 29.12 (22.07, 36.61)* | 32.63 (24.15, 41.75)* | 42.93 (34.03, 52.00)* |
| | SegMamba | 0.29 (0.24, 0.33)* | 0.49 (0.46, 0.53)* | 0.27 (0.22, 0.32)* | 12.39 (10.29, 15.32)* | 7.86 (6.50, 9.61)* | 10.70 (8.71, 13.24)* |
| | DSM | 0.36 (0.32, 0.40)* | 0.58 (0.55, 0.61)* | 0.36 (0.30, 0.41)* | 8.35 (7.50, 9.24)* | 5.32 (4.75, 5.94)* | 8.33 (6.85, 10.07)* |
| | MambaX-Net | **0.76 (0.74, 0.78)** | **0.82 (0.79, 0.84)** | **0.60 (0.54, 0.66)** | **2.76 (2.46, 3.08)** | **1.90 (1.54, 2.31)** | **4.37 (3.49, 5.63)** |

Panel B. Performance by zone-specific ground-truth volume tertile

| Zone | Model | DSC ↑ | | | HD95 (mm) ↓ | | |
|---|---|---|---|---|---|---|---|
| | | Small | Medium | Large | Small | Medium | Large |
| WP | nnU-Net | 0.83 (0.82, 0.85)* | 0.85 (0.83, 0.86)* | 0.84 (0.81, 0.86)* | 8.82 (5.42, 13.13)* | 10.97 (6.11, 17.23)* | 6.52 (5.39, 7.91)* |
| | SwinUNETR | 0.50 (0.35, 0.64)* | 0.26 (0.10, 0.42)* | 0.24 (0.09, 0.40)* | 48.73 (32.94, 65.78)* | 68.01 (49.69, 85.32)* | 67.28 (49.80, 83.21)* |
| | SegMamba | 0.72 (0.66, 0.77)* | 0.76 (0.72, 0.80)* | 0.79 (0.75, 0.82)* | 29.51 (18.93, 41.29)* | 22.75 (12.86, 34.62)* | 15.12 (8.22, 24.36)* |
| | DSM | 0.78 (0.75, 0.81)* | 0.81 (0.80, 0.83)* | 0.80 (0.77, 0.82)* | 10.39 (6.53, 15.32)* | 10.44 (6.10, 16.60)* | 7.34 (6.42, 8.44)* |
| | MambaX-Net | **0.88 (0.87, 0.90)** | **0.91 (0.90, 0.92)** | **0.90 (0.89, 0.91)** | **7.25 (4.11, 11.72)** | **3.67 (3.14, 4.44)** | **4.67 (3.93, 5.51)** |
| TZ | nnU-Net | 0.79 (0.76, 0.81)* | 0.84 (0.82, 0.86)* | 0.81 (0.78, 0.84)* | 11.43 (5.99, 19.84)* | 7.48 (4.23, 12.13) | 7.69 (5.58, 10.36)* |
| | SwinUNETR | 0.45 (0.29, 0.59)* | 0.28 (0.12, 0.45)* | 0.22 (0.08, 0.38)* | 44.85 (27.58, 64.07)* | 61.70 (43.37, 79.55)* | 64.41 (47.64, 80.33)* |
| | SegMamba | 0.67 (0.62, 0.72)* | 0.76 (0.70, 0.80)* | 0.76 (0.71, 0.80)* | 15.10 (10.85, 19.83)* | 15.39 (7.92, 25.66)* | 9.71 (6.58, 14.39)* |
| | DSM | 0.75 (0.71, 0.78)* | 0.82 (0.80, 0.84)* | 0.80 (0.77, 0.82)* | 9.93 (6.62, 14.03)* | 9.57 (5.22, 15.82)* | 6.44 (5.52, 7.57)* |
| | MambaX-Net | **0.84 (0.82, 0.85)** | **0.89 (0.87, 0.90)** | **0.88 (0.87, 0.89)** | **6.09 (4.15, 9.14)** | **6.86 (3.88, 10.99)** | **4.44 (3.82, 5.22)** |
| PZ | nnU-Net | 0.60 (0.55, 0.64)* | 0.56 (0.51, 0.61)* | 0.58 (0.53, 0.63)* | 9.27 (6.94, 12.80)* | 9.02 (7.75, 10.28)* | 10.34 (8.83, 11.94)* |
| | SwinUNETR | 0.28 (0.17, 0.39)* | 0.16 (0.07, 0.26)* | 0.11 (0.03, 0.21)* | 33.14 (21.33, 45.62)* | 47.25 (33.81, 61.55)* | 61.57 (47.06, 75.63)* |
| | SegMamba | 0.41 (0.35, 0.48)* | 0.37 (0.31, 0.42)* | 0.43 (0.38, 0.49)* | 23.20 (15.10, 32.29)* | 31.98 (21.60, 43.69)* | 12.34 (10.89, 13.97)* |
| | DSM | 0.53 (0.48, 0.58)* | 0.46 (0.43, 0.50)* | 0.50 (0.46, 0.54)* | 9.48 (7.20, 13.11)* | 9.34 (8.00, 10.73)* | 10.51 (9.31, 11.88)* |
| | MambaX-Net | **0.75 (0.71, 0.78)** | **0.76 (0.73, 0.79)** | **0.80 (0.76, 0.82)** | **5.72 (3.74, 9.08)** | **4.67 (3.83, 5.66)** | **4.99 (4.06, 6.01)** |

*Note.* Zone-specific tertile thresholds were WP: small = 24.72–42.55 mL, medium = 42.61–68.54 mL, large = 68.76–171.21 mL; TZ: small = 10.84–27.84 mL, medium = 28.95–45.65 mL, large = 46.34–157.20 mL; PZ: small = 7.82–14.71 mL, medium = 15.01–19.45 mL, large = 20.43–36.73 mL.

* Indicates a statistically significant difference ($\alpha < 0.05$) from the best result within the same group.

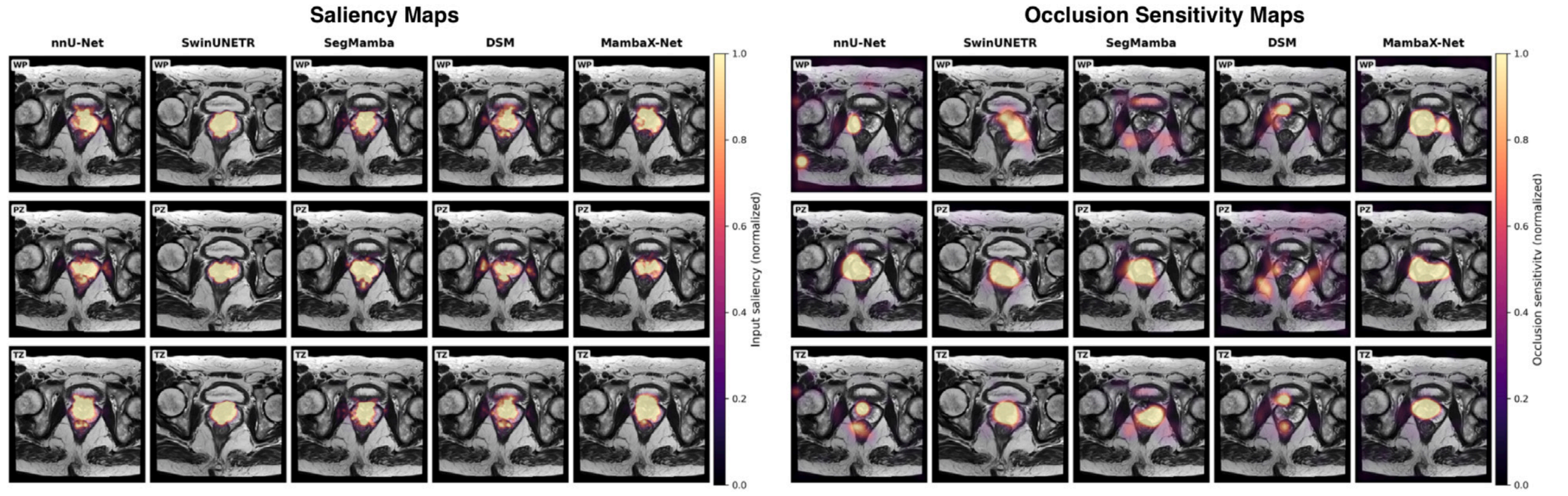

**Fig. 6.** Saliency and occlusion sensitivity maps visualised for all models fine-tuned on $AS_{train}$ with $n = 50$ and tested on a single patient from $AS_{test}$ at t = 1. The three rows in each plot correspond to WP, PZ and TZ as shown by the labels.

(Tables 1–3) and in qualitative examples (Fig. 4), where MambaX-Net produces fewer segmentation errors at the prostate apex and base.

Table 4 provides a deeper analysis of MambaX-Net's performance for $n = 50$ at $t = 1$ on prostate sub-regions and volume-stratified analyses. Panel A shows that MambaX-Net achieved the highest DSC and lowest HD95 across all anatomical sub-regions (apex, mid-gland, and base) for every zone, yielding statistically significant improvements over baseline models in all comparisons. Segmentation quality peaked at the mid-gland and proved most challenging at the apex; nevertheless, MambaX-Net maintained a substantial advantage in this region, particularly in the PZ. Panel B demonstrates that this superiority remained consistent across all prostate volume tertiles (small, medium, large), confirming MambaX-Net's robustness to inter-patient anatomical variability in both gland size and zone-specific volumes.

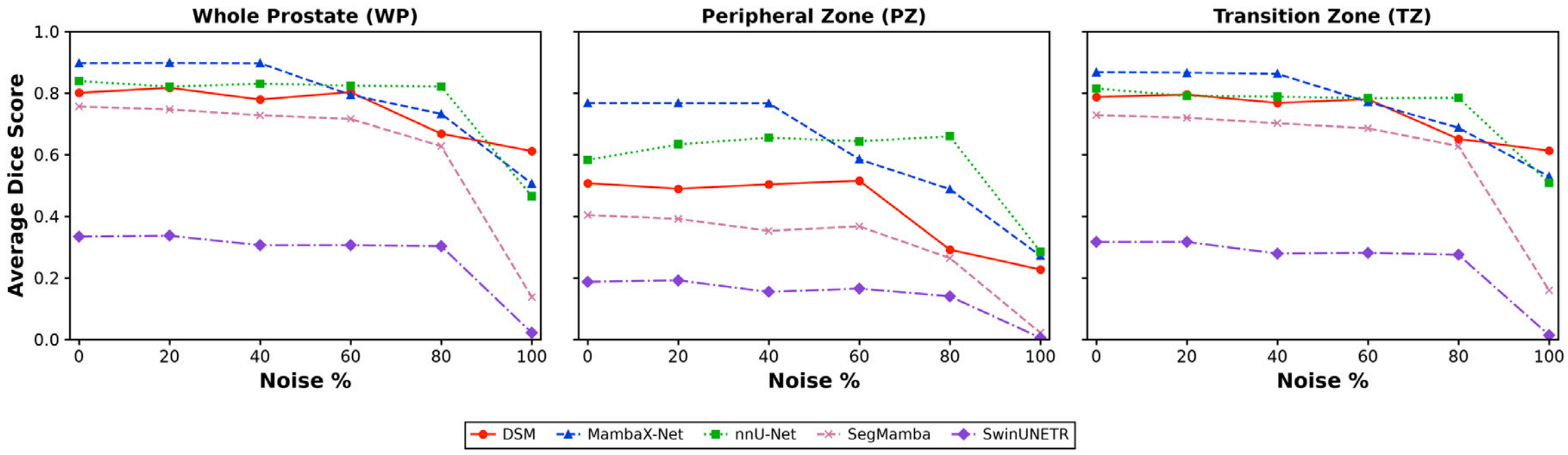

**Fig. 7.** Effect of pseudo-label noise on the model's performance stratified by prostate zone. Noise % represents the percentage of perturbed pseudo-labels used for training. The $y$-axis shows the mean DSC computed on $AS_{test}$ at t = 1.

To provide a more rigorous explainability analysis, we evaluated input-gradient saliency maps and occlusion sensitivity heatmaps (Fig. 6). These analyses estimate the image regions that most strongly influence the network's segmentation output. Across all evaluated models, the attribution maps were anatomically coherent, focusing primarily on gland boundaries and zonal interfaces rather than spurious background regions. However, the models differ significantly in their spatial precision. While nnU-Net and SwinUNETR showed attribution leakage into anterior peri-prostatic tissue, MambaX-Net's response remains more compact and anatomically localised. We quantified this spatial concentration by measuring the fraction of saliency mass located within the ground-truth zones and their immediate boundary bands. MambaX-Net showed the highest spatial concentration, with an average saliency mass of 0.21 inside the target zone and 0.43 within the boundary band (rising to 0.31 in-zone for the whole prostate). These values exceeded the strongest baseline, nnU-Net, which achieved 0.17 in-zone and 0.35 within the boundary band, and remained higher than the DSM, which achieved 0.11 and 0.23, respectively. By localising its predictive evidence more strictly within the target anatomy, MambaX-Net provides an interpretable visual rationale for its improved segmentation performance.

## 4.2. Impact of pseudo-labels quality

In this subsection, we assess how pseudo-label quality influences models' accuracy. We quantified model robustness to label noise by injecting controlled levels of random noise into the pseudo-labels used for training.

Given the counter-intuitive finding in Table 1 that MambaX-Net and DSM performance degraded when a larger number of training patients ($n \geq 100$) were used, we investigated whether this behaviour was driven by the accumulation of noise in the pseudo-labels. While a small number acts as a form of regularisation, a larger volume would cause overfitting to incorrect supervision and catastrophic forgetting of the model's valuable pre-trained knowledge. To examine this effect, we trained the models using 100 patients from $AS_{train}$ by incrementally degrading a percentage of the pseudo-labels with perturbation by eroding the segmentation masks.

Fig. 7 compares the mean DSC computed on $AS_{test}$ across models at increasing levels of training pseudo-label noise, with results stratified by WP, PZ, and TZ. For all models, the zone-stratified curves show that the effect of pseudo-label noise is not uniform across the prostate zones. Performance in the WP and TZ is comparatively stable up to moderate levels of noise, consistent with their larger volumes and more spatially compact morphology. In contrast, the PZ is the most noise-sensitive subregion, consistent with its thin, irregular anatomy and its strong dependence on accurate boundary localisation for delineation.

Single-scan models show greater tolerance to noisy labels in the WP, PZ and TZ, degrading only after introducing 80% noise into the pseudo-labels, which demonstrates higher robustness than the dual-scan models. However, SegMamba and SwinUNETR remain substantially below the other models in terms of performance, especially in the PZ across noise levels and collapse under complete label corruption. Compared to single-scan models, DSM is more sensitive to noise than nnU-Net, although they use the same number of pseudo-labels during training. This increased sensitivity may be attributable to DSM processing two image inputs, both of which can contain noise or image quality degradation.

For MambaX-Net, PZ performance remains high up to 40% pseudo-label noise but decreases earlier than in the WP and TZ once the injected noise level reaches approximately 60%. Under low-to-moderate pseudo-label noise, MambaX-Net achieves the highest DSC in WP, PZ, and TZ, showing that SEM and M-CAM use the prior mask effectively when it remains anatomically reliable. However, when a large fraction of pseudo-labels is corrupted, the same prior can transmit incorrect zonal shape information, making the PZ particularly vulnerable to error propagation.

## 4.3. MambaX-Net backbone selection

To identify the best pre-trained backbone for MambaX-Net, we compared three models, namely, nnU-Net$_{\mathrm{PICAI}}$, SwinUNETR$_{\mathrm{PICAI}}$, and SegMamba$_{\mathrm{PICAI}}$, trained on $PICAI_{train}$ and tested on $PICAI_{Test}$. As shown in Table 5, nnU-Net$_{\mathrm{PICAI}}$ achieved the best performance for both DSC and HD95, consistent with previously reported SOTA results for this benchmark (mean DSC values of 0.90, 0.79, and 0.87 for WP, PZ, and TZ, respectively) (Fassia et al., 2024). Statistical tests confirmed that nnU-Net$_{\mathrm{PICAI}}$ outperformed both SwinUNETR$_{\mathrm{PICAI}}$ and SegMamba$_{\mathrm{PICAI}}$ with statistical significance across all the metrics. Accordingly, nnU-Net$_{\mathrm{PICAI}}$ was selected as the backbone for MambaX-Net.

## 4.4. Ablation study and model efficiency analysis

We performed an ablation study on 26 patients (52 scans) from $AS_{val}$ to assess the contributions of the SEM and Mamba blocks, and to evaluate whether explicit registration between time points influences model performance. We acknowledge that 52 scans is not a big sample size for ablation studies, but given the difficult and time-consuming nature of manual labelling, most of our limited expert annotations were reserved for testing the models. Results in Table 6 demonstrate that while adding either component individually improves performance, combining them yields the best results, confirming their complementary benefits. Moreover, pre-registering multi-time-point T2W scans of the same patient before inputting them into MambaX-Net provided no measurable performance benefit.

Table 7 highlights the trade-off between MambaX-Net's static size (426.4M parameters) and its practical deployment efficiency relative to SOTA baseline models. Although MambaX-Net is the largest model

**Table 5**
Segmentation performance of the baseline models on $PICAI_{Test}$ was evaluated using DSC and HD95 across the WP, PZ, and TZ. The best model is highlighted in **bold**.

| Model | DSC ↑ | | | HD95 (mm) ↓ | | |
|---|---|---|---|---|---|---|
| | WP | PZ | TZ | WP | PZ | TZ |
| nnU-Net$_{PICAI}$ | **0.94 ± 0.02** | **0.85 ± 0.06** | **0.91 ± 0.04** | **3.1 ± 1.5** | **3.3 ± 2.4** | **3.4 ± 1.9** |
| SwinUNETR$_{PICAI}$ | 0.92 ± 0.02* | 0.81 ± 0.06* | 0.88 ± 0.05* | 5.5 ± 2.1* | 5.8 ± 3.1* | 6.0 ± 2.3* |
| SegMamba$_{PICAI}$ | 0.91 ± 0.03* | 0.79 ± 0.07* | 0.88 ± 0.05* | 4.0 ± 1.7* | 5.0 ± 3.4* | 4.6 ± 1.9* |

* Indicating a statistically significant difference ($\alpha < 0.05$) from the best value.

**Table 6**
Ablation results of MambaX-Net computed on the $AS_{val}$ data at $t = 1$ with DSC reported for WP, PZ, and TZ of the prostate.

| Registration | SEM | Mamba | WP | PZ | TZ |
|---|---|---|---|---|---|
| ✗ | ✗ | ✗ | 0.69 ± 0.10 | 0.51 ± 0.15 | 0.67 ± 0.12 |
| ✗ | ✓ | ✗ | 0.75 ± 0.08 | 0.60 ± 0.11 | 0.72 ± 0.09 |
| ✗ | ✗ | ✓ | 0.81 ± 0.06 | 0.64 ± 0.08 | 0.77 ± 0.07 |
| ✗ | ✓ | ✓ | 0.88 ± 0.06 | 0.76 ± 0.09 | 0.85 ± 0.06 |
| ✓ | ✓ | ✓ | 0.87 ± 0.08 | 0.75 ± 0.10 | 0.86 ± 0.06 |

**Table 7**
Model efficiency analysis on an Nvidia A100 40 GB GPU. *Inf T* is the inference time, *#Params* the number of model parameters measured in millions, and *Mem* the peak memory in GB used for a single forward and backwards pass.

| Model | GFLOPs | #Params | Mem (GB) | Inf T (ms) |
|---|---|---|---|---|
| nnU-Net | 1862.3 | 44.6 | 6.6 | 49.3 |
| SwinUNETR | 1888.7 | 72.6 | 25.1 | 649.5 |
| SegMamba | 1756.2 | 66.9 | 14.2 | 223.3 |
| DSM | 2738.6 | 44.6 | 14.3 | 928.8 |
| MambaX-Net | 2953.1 | 426.4 | 14.4 | 276.3 |

by parameter count, this primarily affects checkpoint storage rather than inference feasibility. Because the M-CAM layers scale linearly, MambaX-Net avoids the quadratic memory bottlenecks of Transformer-based models like SwinUNETR. Consequently, MambaX-Net achieves the highest overall segmentation accuracy while requiring substantially less peak GPU memory than SwinUNETR (14.4 GB vs. 25.1 GB) and operating 2.4 times faster than SwinUNETR (276.3 ms vs. 649.5 ms). While the single-scan nnU-Net is inherently lighter (6.6 GB) and faster (49.3 ms), MambaX-Net's superior boundary delineation and longitudinal tracking support this additional computational cost.

For clinical deployment, the 14.4 GB peak memory footprint allows MambaX-Net to run on standard mid-range hospital GPUs (e.g., NVIDIA RTX 4080, 16 GB VRAM) without requiring input downsampling, which typically degrades accuracy. However, deployment may remain challenging for sites operating on smaller GPUs, for example, below 12 GB VRAM, due to the dual-encoder structure required to process two temporal scans. To accommodate tighter hardware constraints in future work, sequential encoder inference with cached features could halve activation memory. Furthermore, deploying the model with half-precision (FP16) inference or INT8 quantisation, structural pruning of the encoder channels, and knowledge distillation provide feasible compression strategies for optimising MambaX-Net in resource-limited clinical settings.

## 5. Conclusion

In this work, we proposed MambaX-Net, a novel dual-scan architecture designed to integrate longitudinal information that leverages pseudo-labels from a pre-trained nnU-Net within a self-training framework. MambaX-Net significantly outperformed SOTA models for prostate MRI segmentation through the integration of the newly proposed Mamba-enhanced Cross-Attention Module (M-CAM) and Shape Extractor Module (SEM). These components enable the model to integrate longitudinal information and effectively adapt to new domains using only noisy pseudo-labels, demonstrating a feasible path towards clinical translation without manual annotation or explicit image registration. Our experiments showed that, even when trained with a small dataset, MambaX-Net achieved higher performance than nnU-Net or SwinUNETR trained on a larger dataset. Moreover, our investigation revealed that the performance of MambaX-Net can degrade more rapidly than that of single-scan models as the volume of noisy pseudo-labels increases.

This study has several limitations. First, validation was performed on a single-institution AS dataset of limited size, and used for MambaX-Net development and optimisation. However, because the expert-annotated cohort contained highly challenging cases (e.g., poor boundary delineation, low SNR, and TURP-treated patients), this serves as a rigorous stress test under the hardest realistic conditions of a clinical setting with real-world data. Consequently, the reported metrics are conservative and more likely to underestimate than inflate the model's generalisability to typical AS cases. Second, MambaX-Net is explicitly designed as a dual-input model and does not involve joint modelling of three or more time points. The experiments at $t = 2$ and $t = 3$ therefore demonstrate rolling pairwise inference across subsequent visits, using the same model weights as at $t = 1$. In long-term monitoring, this design may accumulate prediction drift if the previous mask input to MambaX-Net is itself a model prediction, particularly for noise-sensitive prostate regions such as the PZ. Clinical deployment should therefore include uncertainty-based quality control and periodic anchoring to radiologist-approved masks, when available. Third, the varying quality of the pseudo-labels was not explicitly accounted for in MambaX-Net's learning process. Future work will focus on addressing these constraints to ensure robust clinical deployment. External multi-centre validation remains essential, particularly to characterise the model's behaviour across different scanner vendors, magnetic field strengths, and acquisition protocols. Promising strategies for mitigating pseudo-label noise include uncertainty-aware self-training, such as UAMT-style consistency between teacher and student segmentations, confidence-weighted pseudo-label selection so that only voxels with high backbone confidence contribute to the loss, and consistency regularisation across augmentations or across the two paired time points. Based on the zone-stratified noise analysis, these mitigation strategies are likely to be particularly beneficial for the PZ, where label noise has the greatest impact on segmentation performance, especially along its boundaries. We view these as natural next steps for improving robustness without requiring additional expert annotation. Finally, we aim to extend its application to other longitudinal tasks, such as tracking prostate lesion progression in AS patients.

## CRediT authorship contribution statement

**Yovin Yahathugoda:** Writing – original draft, Methodology, Investigation, Conceptualization. **Davide Prezzi:** Data curation. **Patricia A. Gutierrez:** Data curation. **Piyalitt Ittichaiwong:** Data curation. **Vicky Goh:** Supervision, Data curation. **Sebastien Ourselin:** Supervision. **Michela Antonelli:** Writing – review & editing, Supervision, Project administration, Conceptualization.

## Code availability

The implementation of MambaX-Net will be made publicly available at: https://github.com/yoviny/MambaX-Net.git

## Declaration of competing interest

The authors declare the following financial interests/personal relationships which may be considered as potential competing interests: Yovin Ransika Yahathugoda reports financial support was provided by Engineering and Physical Sciences Research Council. Michela Antonelli reports financial support was provided by Prostate Cancer UK. Given his role as Associate Editor, Prof. Sebastien Ourselin had no involvement in the peer review of this article and had no access to information regarding its peer review. Full responsibility for the editorial process for this article was delegated to another journal editor. If there are other authors, they declare that they have no known competing financial interests or personal relationships that could have appeared to influence the work reported in this paper. If there are other authors, they declare that they have no known competing financial interests or personal relationships that could have appeared to influence the work reported in this paper.

## Acknowledgements

This work was funded by the Engineering & Physical Sciences Research Council (EPSRC), part of the EPSRC DTP, Grant Ref: EP/W524475/1, and supported by Prostate Cancer UK through a Research Innovation Award, Grant Ref: RIA24-ST2-015.

## Appendix A. Supplementary data

Supplementary material related to this article can be found online at https://doi.org/10.1016/j.media.2026.104251.

## Data availability

The PI-CAI dataset and ProstateX subset are publicly available. The in-house AS data cannot be shared due to patient privacy governance.